%% file: main.tex
\documentclass{article}

\usepackage[utf8]{inputenc}
\usepackage[T1]{fontenc}
\usepackage{main}
\usepackage{microtype}
\usepackage{graphicx}
\usepackage{times}
\usepackage{inconsolata}
\usepackage{amsmath}
\usepackage{amssymb}
\usepackage{enumitem}
\usepackage{booktabs}
\usepackage{textcomp}
\usepackage{xspace}
\usepackage{float}
\usepackage{multirow}
\usepackage{fancyhdr}
\usepackage{wrapfig}
\usepackage{algorithm}
\usepackage[noend]{algpseudocode}

\definecolor{mydarkblue}{rgb}{0,0.08,0.45}
\usepackage[colorlinks=true,linkcolor=mydarkblue,citecolor=mydarkblue,filecolor=mydarkblue,urlcolor=mydarkblue]{hyperref}
\setlist[itemize,1]{leftmargin=10pt}

\input{commands}

\definecolor{YaleBlue}{RGB}{0, 53, 107}
\definecolor{UChiRed}{RGB}{128, 0, 0}
\definecolor{UIUCO}{RGB}{255, 95, 5}
\definecolor{TCSC}{RGB}{1, 126, 199}

\newcommand{\UChi}{\hspace{.1em}^{\textcolor{UChiRed}{\boldsymbol{C}}}}
\newcommand{\Yale}{\hspace{.1em}^{\textcolor{YaleBlue}{\boldsymbol{Y}}}}
\newcommand{\UIUC}{\hspace{.1em}^{\textcolor{UIUCO}{\boldsymbol{I}}}}
\newcommand{\TCS}{\hspace{.1em}^{\textcolor{TCSC}{\boldsymbol{T}}}}

\newcommand{\huggingface}{\raisebox{-1.5pt}{\includegraphics[height=1.05em]{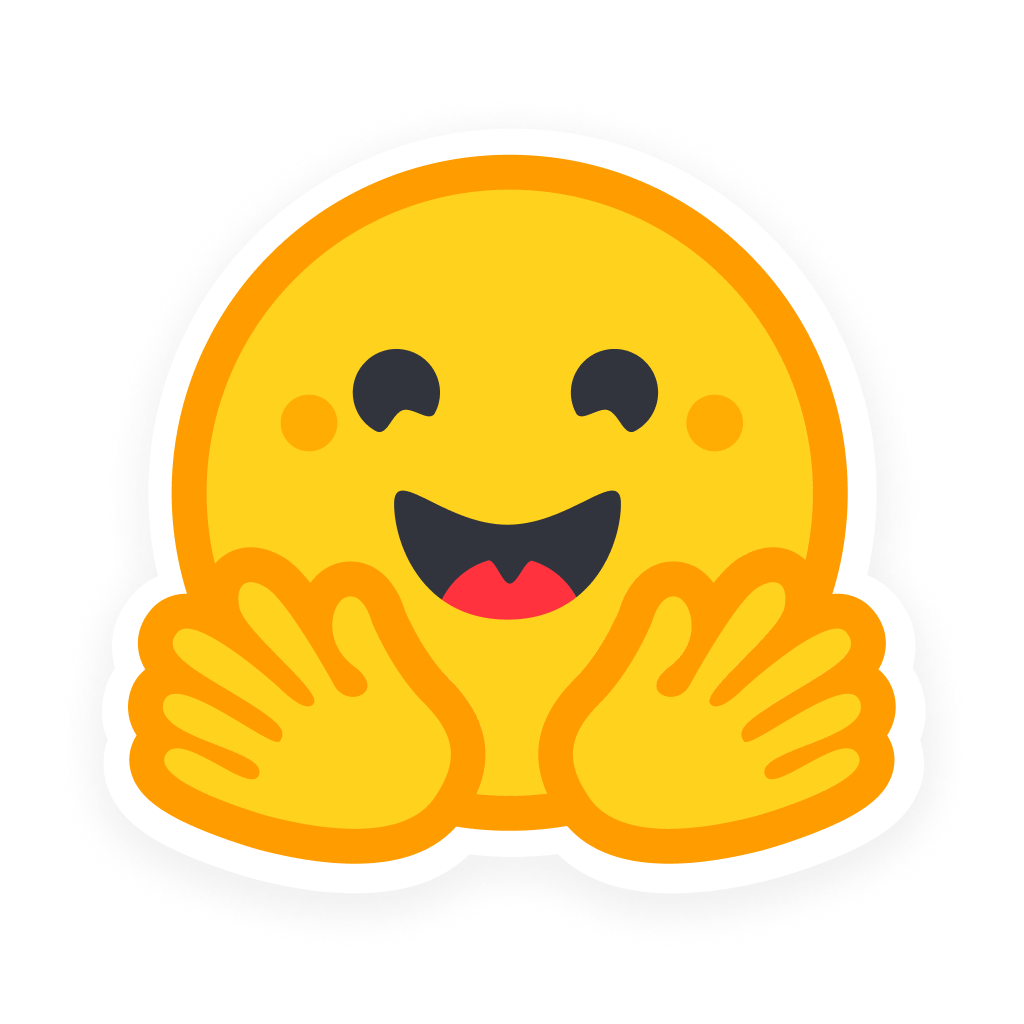}}\xspace}
\newcommand{\github}{\raisebox{-1.5pt}{\includegraphics[height=1.05em]{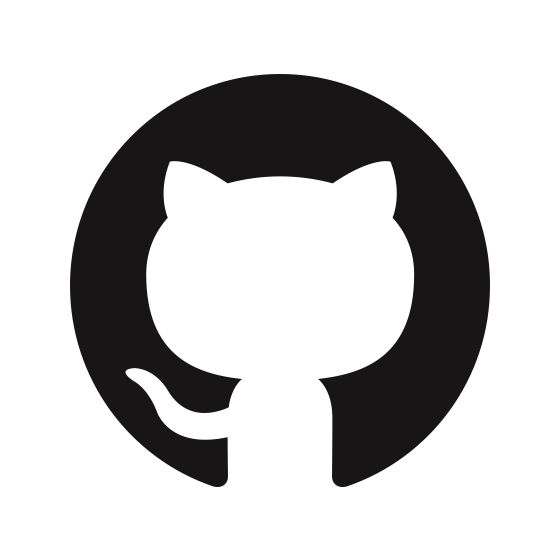}}\xspace}
\newcommand{\anchoricon}{\raisebox{-1.5pt}{\includegraphics[height=1.0em]{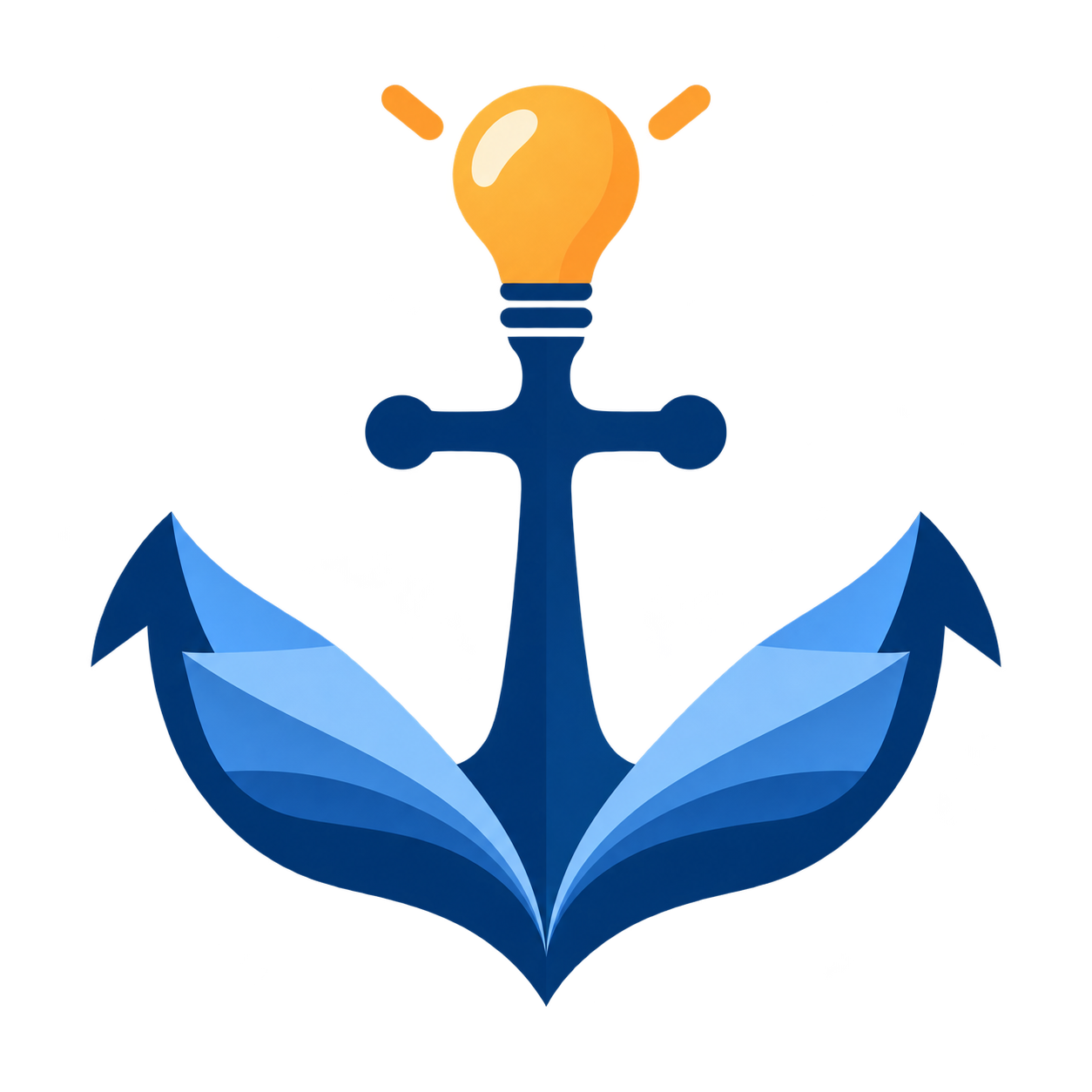}}\xspace}
\newcommand{\anchorlogo}{\raisebox{-0.25em}{\includegraphics[height=1.3em]{logo/ideaanchor.png}}}

\definecolor{goalblue}{RGB}{31,78,121}
\definecolor{goalbluebg}{RGB}{232,240,250}
\definecolor{guidepurple}{RGB}{139,30,63}
\definecolor{guidepurplebg}{RGB}{250,235,240}
\definecolor{rubricgreen}{RGB}{45,106,79}
\definecolor{rubricgreenbg}{RGB}{235,247,241}
\definecolor{outputamber}{RGB}{150,100,30}
\definecolor{outputamberbg}{RGB}{255,249,235}
\definecolor{thinkColor}{HTML}{2A7AB5}
\definecolor{solColor}{HTML}{2AA587}
\definecolor{rubricColor}{HTML}{7570b3}
\definecolor{anchorColor}{HTML}{D4740A}

\newcommand{\anchor}{\texttt{IdeaAnchor}\xspace}

\title{\fontsize{15}{17}\selectfont \anchorlogo~\anchor: Teaching LLMs to Turn Literature into Research Ideas}

\author{
\textbf{Ziyu Chen}$\UChi$ \qquad
\textbf{Yilun Zhao}$\Yale$ \qquad
\textbf{Jiashuo Sun}$\UIUC$ \qquad
\textbf{Yiling Ma}$\Yale$ \\ [3pt]
\textbf{Manasi Patwardhan}$\TCS$ \qquad
\textbf{Arman Cohan}$\Yale$ \\ [7pt]
$\Yale$Yale University \qquad $\UChi$The Unversity of Chicago \\ [3pt]
$\UIUC$University of Illinois Urbana-Champaign \qquad
$\TCS$Tata Consultancy Services\\ [8pt]
\href{https://ziyu.ch/research/ideaanchor}{\anchoricon \texttt{Homepage}} \qquad
\href{https://github.com/ziyuuc/IdeaAnchor}{\github \texttt{Code}} \qquad
\href{https://huggingface.co/datasets/idealand/IdeaAnchor}{\huggingface \texttt{Dataset}}\\[4pt]
}

\begin{document}

\maketitle
\thispagestyle{fancy}
\fancyhead{}
\lhead{%
    \includegraphics[height=1.1cm]{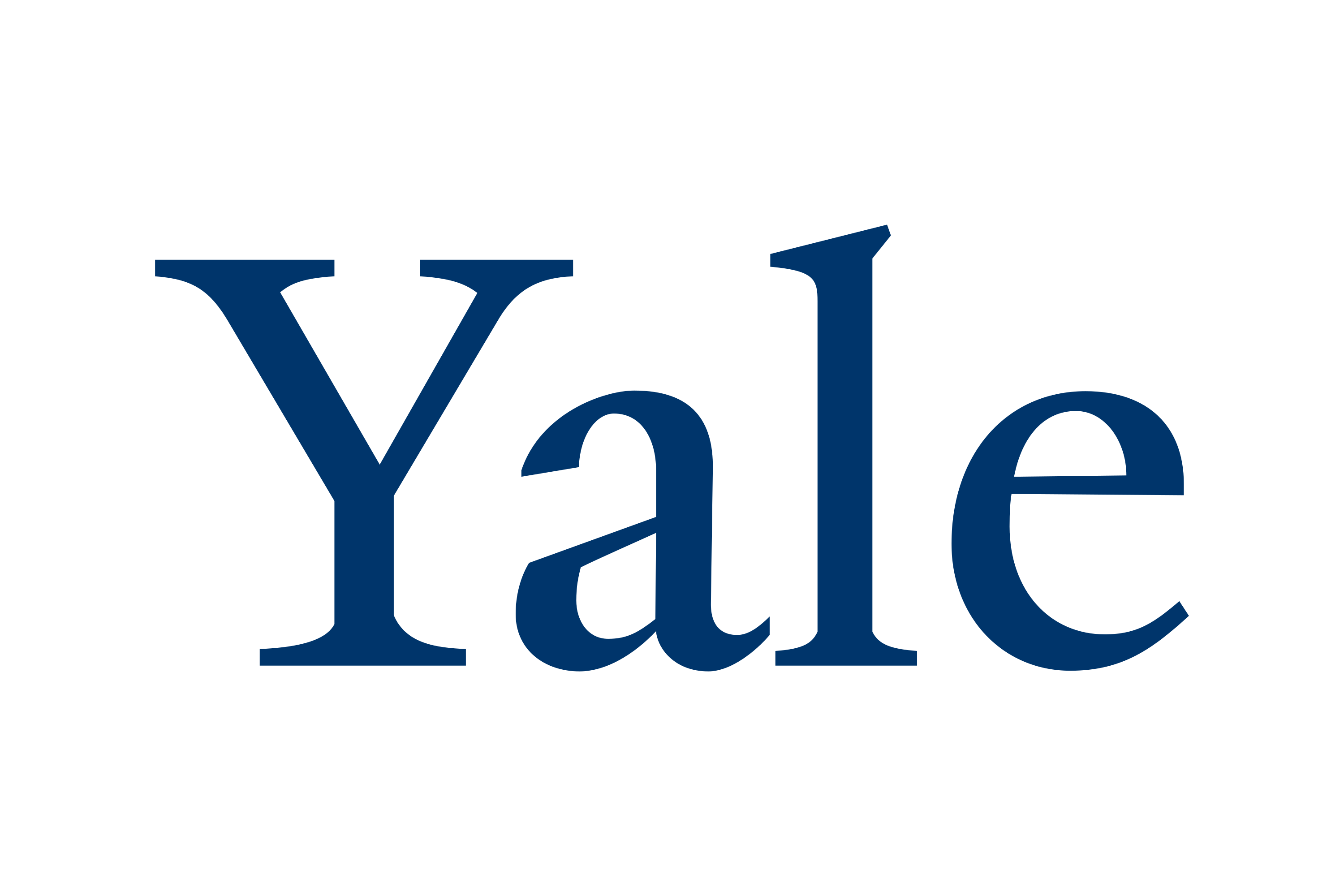}\hspace{0.2cm}%
    \includegraphics[height=1.05cm]{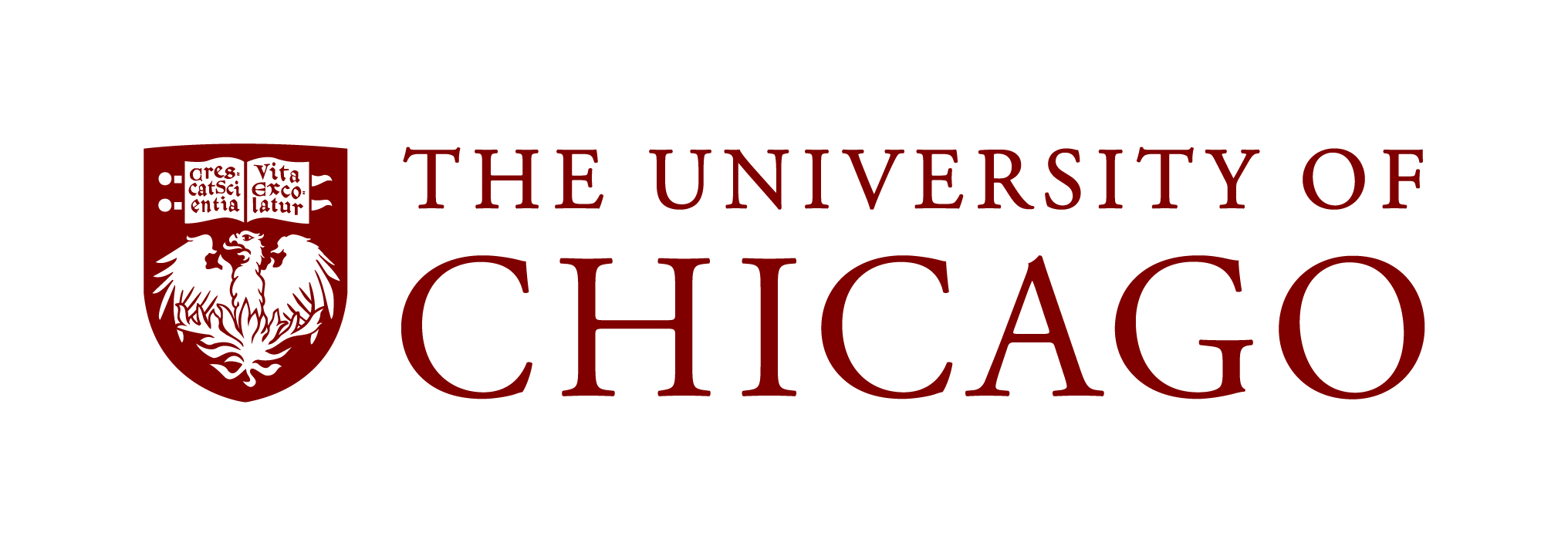}\hspace{0.3cm}%
    \raisebox{0.25cm}{\includegraphics[height=0.57cm]{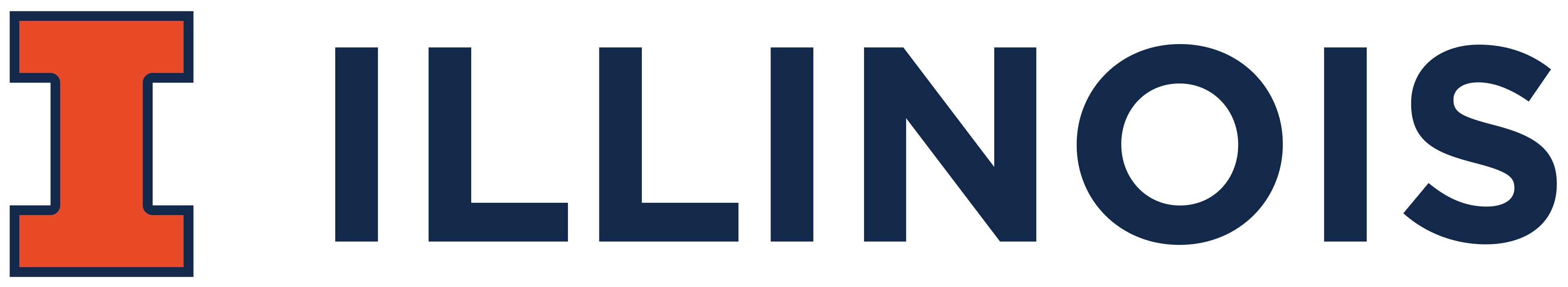}\hspace{0.36cm}}%
    \raisebox{0.05cm}{\includegraphics[height=0.8cm]{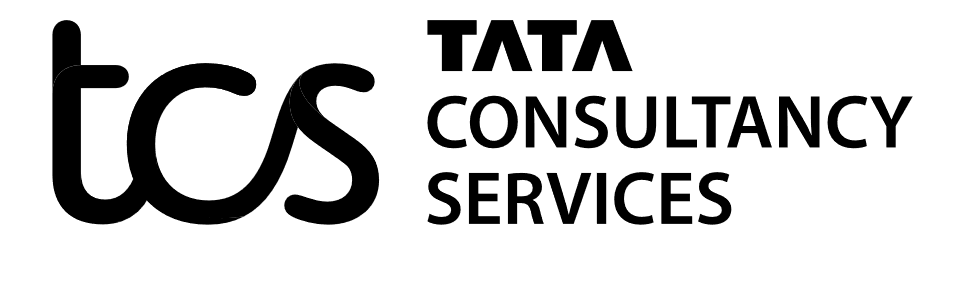}}%
}

\vspace{0.5em}

\fancyfoot[C]{\thepage}
\renewcommand{\headrulewidth}{0pt}
\setlength{\headheight}{12pt}
\addtolength{\topmargin}{0pt}
\setlength{\headsep}{3mm}

\vspace{-0.5em}

\pagestyle{plain}
\setlength{\parskip}{0.5ex plus 0.2ex minus 0.1ex}

\input{section/0-abstract}

\input{section/1-intro}
\input{section/2-anchor}
\input{section/3-method}
\input{section/4-exp}
\input{section/5-related}
\input{section/6-conclusion}

\newpage

\bibliographystyle{unsrtnat}
\bibliography{main}

\appendix

\newpage

\input{section/X-suppl}

\end{document}

%% file: commands.tex
\usepackage{tikz}
\usepackage{xcolor}
\usetikzlibrary{arrows.meta}
\definecolor{TaxBlueBg}{HTML}{EDF4FA}
\definecolor{TaxBlueHead}{HTML}{D7E8F7}
\definecolor{TaxBlueLine}{HTML}{5C8DB8}
\definecolor{TaxGreenBg}{HTML}{EEF7F1}
\definecolor{TaxGreenHead}{HTML}{D8ECD9}
\definecolor{TaxGreenLine}{HTML}{6CA47A}
\definecolor{TaxRedBg}{HTML}{F9ECEC}
\definecolor{TaxRedHead}{HTML}{F2D6D6}
\definecolor{TaxRedLine}{HTML}{C78383}
\definecolor{TaxSlate}{HTML}{667085}
\usepackage{soul}
\usepackage{amssymb}
\usepackage{amsmath}
\usepackage{xspace}
\usepackage{float}
\usepackage{booktabs}
\usepackage{colortbl}
\usepackage{subcaption}
\usepackage{caption}
\usepackage{bbm}
\usepackage{adjustbox}
\usepackage{makecell}
\usepackage[framemethod=TikZ]{mdframed}

\definecolor{mypositive}{RGB}{0, 128, 0}
\definecolor{mynegative}{RGB}{220, 20, 60}
\definecolor{mypositive}{RGB}{24, 103, 173}
\definecolor{mynegative}{RGB}{255, 128, 0}

\newcommand{\eg}{e.g.,\xspace}

\definecolor{customblue}{HTML}{1F77B4}
\definecolor{customorange}{HTML}{FF7F0E}
\definecolor{customgreen}{HTML}{2CA02C}
\definecolor{no}{HTML}{BA8E23}

\definecolor{darkgreen}{RGB}{0,100,0}
\definecolor{darkred}{RGB}{139,0,0}

\newcommand{\deltaScore}[1]{%
\begingroup
\ifdim #1pt > 0pt
\textcolor{darkred}{\scriptsize$^{#1}$}%
\else
\textcolor{darkgreen}{\scriptsize$^{#1}$}%
\fi
\endgroup
}

\usepackage[skins,breakable]{tcolorbox}

\definecolor{neutralgray}{RGB}{80,80,80}
\definecolor{neutralgraybg}{RGB}{243,243,243}

\NewTColorBox{headerbox}{O{} m m O{}}{%
  enhanced,
  breakable,
  colback=#3,
  colframe=#2,
  colbacktitle=#2,
  coltitle=white,
  fonttitle=\bfseries,
  title=#1,
  arc=2mm,
  boxrule=0.8pt,
  left=3mm,
  right=3mm,
  top=2.5mm,
  bottom=2.5mm,
  toptitle=1mm,
  bottomtitle=1mm,
  lefttitle=3mm,
  righttitle=3mm,
  before skip=6pt,
  after skip=6pt,
  #4,
}


\newtcolorbox{boxblue}{enhanced,colback=blue!5!white,colframe=blue!75!black,breakable=true}

\usepackage{listings}

\lstdefinelanguage{json}{
    basicstyle=\scriptsize\ttfamily,
    numbers=none,
    numberstyle=\tiny,
    stepnumber=1,
    numbersep=5pt,
    showstringspaces=false,
    breaklines=true,
    frame=none,
    backgroundcolor=\color{white},
    literate=
     *{:}{{{\color{black}{:}}}}{1}
      {,}{{{\color{black}{,}}}}{1}
      {\{}{{{\color{black}{\{}}}}{1}
      {\}}{{{\color{black}{\}}}}}{1}
      {[}{{{\color{black}{[}}}}{1}
      {]}{{{\color{black}{]}}}}{1},
}

%% file: section/0-abstract.tex
\begin{abstract}
Scientific research often begins by synthesizing ideas from a set of related papers to identify gaps and formulate new directions. However, training language models to perform this form of literature-grounded ideation remains challenging, as existing approaches based on prompting or feedback lack structured supervision for how papers should be synthesized.
We introduce \anchor, a paradigm for training LLMs to perform research ideation, using structured specifications as privileged signals. Each \anchor instance encodes how each input paper should be synthesized into a successful idea, including their functional roles, relationships, and target synthesis criteria. We build this paradigm by mining instances from published papers, capturing how real ideas emerge from prior literature. We then train models via demonstration, self-distillation, and reinforcement learning, and further enhance generation with retrieval at inference time. Experiments show consistent improvements in ideation quality. Our analysis reveals a functional decomposition: anchor-based training strengthens \textit{creative synthesis}, retrieval enhances \textit{detail elaboration}, and combining both yields the best performance.
\end{abstract}

%% file: section/1-intro.tex
\input{figure_tex/figure1}
\section{Introduction}

Scientific discoveries rarely emerge from a blank page. Before any novel finding, researchers read a cluster of related papers, identify what each one enables, locate what remains unresolved, and crystallize a research direction that is both new and defensible~\citep{uzzi2013atypical,park2023papers,pu2025ideasynth}. This process decomposes naturally into two capabilities. (i) \textit{creative synthesis}: analyzing prior work at a high level, recognizing gaps across multiple contributions, and formulating a research direction that no single paper addresses; (ii) \textit{detail elaboration}: fleshing out the envisioned approach with concrete methodological specifics drawn from the technical substance of prior work. A strong research idea requires both a compelling direction grounded in literature and enough technical depth to be actionable.

LLMs have demonstrated impressive downstream scientific assistance, from summarizing and reasoning over papers~\citep{zhao2025sciarena,chen2026scimdr} to executing well-defined experiments~\citep{boiko2023autonomous,nathani2025mlgym, novikov2025alphaevolvecodingagentscientific,wang2023scientific}. Even so, generating novel and actionable research ideas remains challenging. Given some related papers, LLMs tend to produce fluent overviews or shallow \textit{concept blends}, which often struggle to identify meaningful research gaps and develop concrete methods grounded in existing scholarly evidence~\citep{gupta-pruthi-2025-glitters,si2024llmsgeneratenovelresearch}. Comparing ideas generated from the same prior works, \citet{chen2026measuring} further find a consistent distributional gap from human research taste: LLM ideas concentrate on bridge-like opportunities and synthesis methods, whereas human papers frame gaps and construct contributions in far more diverse ways. Recent efforts address this through goal-conditional plan generation with rubric rewards~\citep{goel2025training}, execution feedback in sandboxed environments~\citep{jansen2025codescientist,lu2024ai}, or single-paper hypothesis inversion~\citep{o2025sparks}. However, these approaches either assume the research goal is already given, require domain-specific execution environments, or do not support multi-paper synthesis.

We argue that the bottleneck of ideation is the absence of training signals that capture \textit{what makes a good research idea emerge from prior work}. To address this, we introduce \anchor, per-instance and literature-derived specifications that encode the functional roles each input paper should play, the gaps that should be identified, and the patterns that should be leveraged in the synthesis. 
\anchor serves as a \textit{unified data interface} for improving research ideation via 3 distinct training paradigms: 
(1) \textbf{Anchor-Guided Demonstration:} By prompting a strong external LLM with the input paper list with corresponding \anchor instance, we can produce high-quality demonstrations that the target model learns from via supervised fine-tuning (SFT). \anchor steer the teacher model to better utilize input literature, yielding higher-quality demonstrations for training.
(2) \textbf{Anchor-Based Self-Distillation:} In the absence of an external teacher, the \textit{target model itself} with \anchor can self-distill~\citep{zhao2026self}. The model's anchor-conditioned outputs, which benefit from seeing the structured specification, serve as self-training targets. This creates a self-improvement loop: the model distills the knowledge embedded in \anchor into its own parameters.
(3) \textbf{Anchor-Privileged Reinforcement Learning:} Privileged information in \anchor can be used for RL reward.~\citep{christiano2017deep,bai2022constitutional}. During GRPO~\citep{shao2024deepseekmathpushinglimitsmathematical}, the reward judge checks each output against anchor-defined criteria. Through iterative reward optimization, the policy model gradually learns to generate contents that align with standards from real publications.

To construct training data for this paradigm at scale, we retrospectively reconstruct how published contributions build on prior work~\citep{jurgens2018measuring,lo2020s2orc}. Specifically, we assemble a corpus of approximately 14K papers spanning machine learning and natural science. Leveraging LLMs, we first extract the core intellectual contribution of each work, then retrospectively trace its most influential prior studies, and further reverse-engineer the implicit anchor linking foundational literature to subsequent research innovations. By exploiting the inherent provenance structure of academic publications, this pipeline produces high-quality structured training signals, and a study with the original authors confirms that the mined prior works closely reflect the literature they credit for their ideas.

To determine whether anchor-derived training signals actually teach models to synthesize research ideas from prior work, evaluation must mirror the same provenance structure. However, existing ideation benchmarks are not designed for this question. For example, \citet{si2024llmsgeneratenovelresearch} evaluate open-ended idea generation with generic rubrics across four fixed dimensions; IdeaBench~\citep{guo2024ideabench} ranks ideas against reference papers but without structured multi-paper input sets; LiveIdeaBench~\citep{ruan2024liveideabench} measures divergent thinking from single-keyword prompts; and SciMON~\citep{wang2024scimon} optimizes novelty given a single background context. These benchmarks typically rate open-ended ideas or single-paper transformations, without expert-verified prior-work sets and instance-specific criteria for multi-paper synthesis. We therefore build a benchmark where domain experts analyze academic papers, pinpoint their pivotal prior studies, and refine standardized evaluation metrics for literature-grounded ideation quality. Our experiments focus on \textit{creative synthesis}: whether a model can recognize non-obvious gaps and formulate novel directions from multiple papers. We evaluate three anchor-driven training paradigms and inference-time retrieval augmentation with both LLM-based judging and human expert assessment~\citep{liu2023g,zheng2023judging}. Across settings, training improves ideation quality, with RL producing the largest gains in creative synthesis; retrieval mainly enhances detailed content elaboration. Our results uncover a functional decomposition: gap recognition and direction formulation need to be internalized via training, whereas grounding conceived ideas into methodological details can be enhanced by inference-time retrieval. 

Our primary contributions can be summarized as follows:

\begin{itemize}[leftmargin=*,itemsep=1pt,topsep=2pt]
    \item We adopt three anchor-driven training paradigms, alongside inference-time retrieval augmentation to boost research ideation capability in LLMs.

    \item We propose a pipeline to mine the intellectual genealogy of academic publications and extract \anchor instances via LLMs, constructing a structured corpus to address the lack of dedicated training signals for research ideation tasks. We also build an expert-annotated benchmark with evaluation rubrics to systematically assess the quality of research ideation from prior works.

    \item Through extensive LLM-based and human evaluation, we verify the effectiveness of our framework and uncover a functional decomposition of research ideation, offering guidance for future research on AI-assisted scientific innovation.

\end{itemize}

%% file: figure_tex/figure1.tex
\begin{figure}[h]
    \centering
    \includegraphics[width=0.9\linewidth]{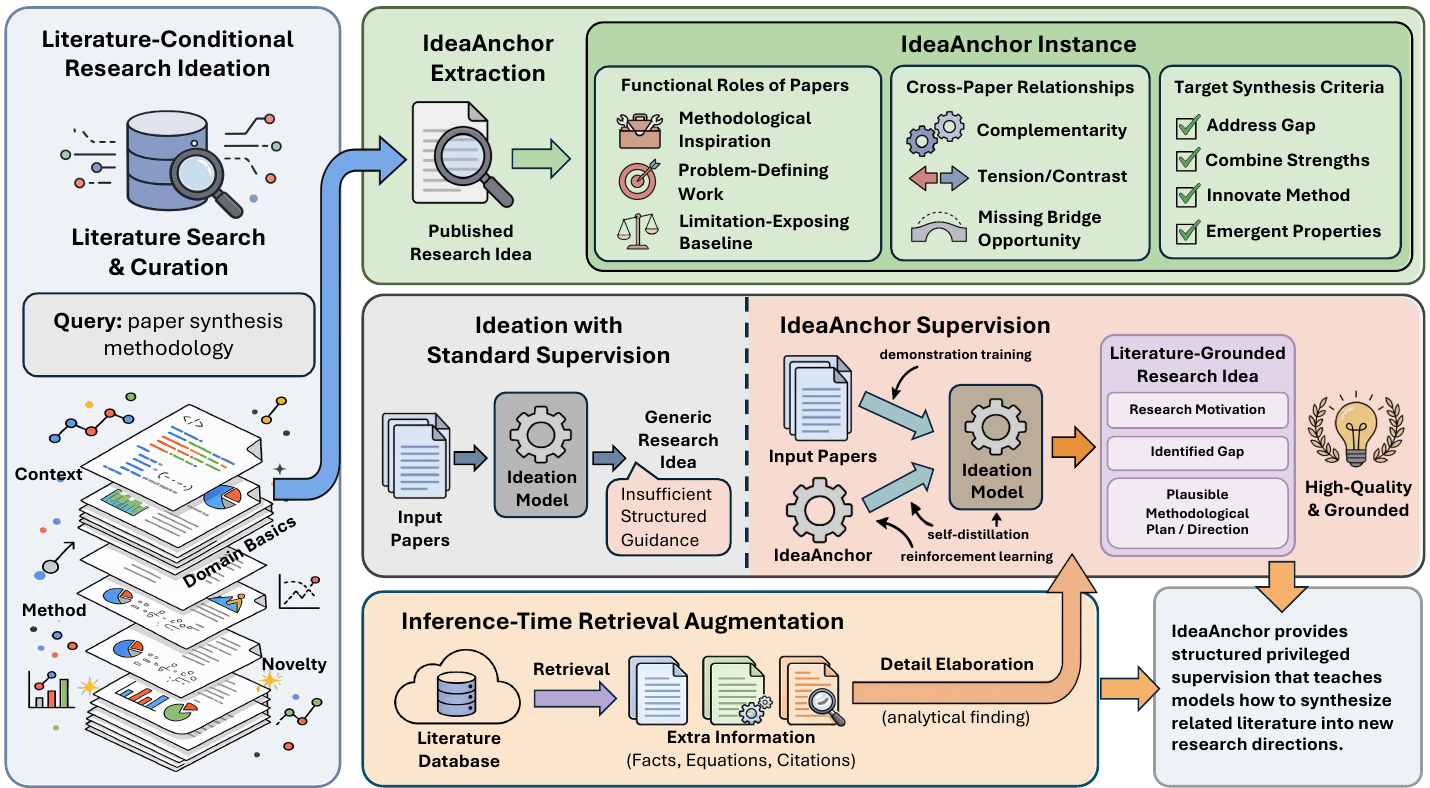}
    \caption{\textbf{Overview of \anchor.}
    Starting from a set of papers, we extract \anchor instances from published papers by identifying the functional roles of prior work, their cross-paper relationships, and the target synthesis criteria. These structured anchors make otherwise implicit synthesis logic explicit, providing privileged supervision for demonstration learning, self-distillation, and reinforcement learning. At inference time, the framework uses either related papers or a topic query as input. It can also incorporate retrieval augmentation, which complements the learned synthesis capability by supplying factual and methodological details for elaborating generated research ideas.}
    \label{fig:overview}
\end{figure}

%% file: section/2-anchor.tex
\section{\anchor}
\label{sec:anchor}

This section first defines the task of \textit{literature-conditional research ideation} and then introduces \anchor, the structured specifications that form the backbone of our training framework.

\subsection{Literature-Conditional Research Ideation}
\label{sec:task}

Let $\mathcal{P} = \{p_1, \ldots, p_N\}$ denote a set of $N$ related papers within a research topic, where each paper $p_i = (t_i, c_i)$ consists of a title $t_i$ and a brief summary $c_i$. A model $\pi_\theta$ is required to produce a structured output $y = (\mathcal{T}, \mathcal{M}, \mathcal{R})$ consisting of:

\begin{itemize}[leftmargin=*,itemsep=1pt,topsep=2pt]
    \item \textbf{Thinking Trace} $\mathcal{T}$: a structured analysis of each input paper's functional role and inter-paper relationships, articulating the synthesis logic that leads to a new research direction.
    \item \textbf{Research Motivation} $\mathcal{M}$: a novel problem definition derived from identifying gaps, limitations, or unexplored combinations across the input papers.
    \item \textbf{Research Plan} $\mathcal{R}$: a detailed methodology that synthesizes techniques from the input papers to address the identified problem.
\end{itemize}

Here, the research motivation $\mathcal{M}$ must be \emph{discovered} from the literature rather than given as input, and the plan $\mathcal{R}$ must be \emph{grounded} in the specific papers provided.

\subsection{\anchor Specification}
\label{sec:anchor_def}

An \textbf{\anchor} $\mathcal{A}_D$ for a training instance is a structured specification derived from a ground-truth paper $D$ and its input literature $\mathcal{P}$. It encodes three types of information:

\begin{itemize}[leftmargin=*,itemsep=2pt,topsep=2pt]
    \item \textbf{Functional Role Assignments} $\{\rho_i\}_{i=1}^N$: Each prior work $p_i$ is assigned one of four roles that characterize its relationship to $D$: \textit{Direct Predecessor} (the method most immediately extended or improved upon), \textit{Inspiration Source} (a technique or idea from a different context that sparked the approach), \textit{Gap \& Motivation} (work whose limitations define the research problem), and \textit{Methodological Ingredient} (a specific technical component incorporated into the proposed method)~\citep{cohan2019structural,jurgens2018measuring}. Each role assignment is accompanied by a \textit{learned-insight} summary: what the authors of $D$ learned from $p_i$ and how it influenced the contribution.

    \item \textbf{Relationship Analysis} $\mathcal{S}_D$: A structured analysis that articulates the synthesis logic connecting the prior works $\mathcal{P}$ to $D$'s contribution:
    \begin{itemize}[leftmargin=*,itemsep=1pt]
        \item \textit{Per-paper analysis}: For each prior work, what it achieved, what limitation remains, and why that limitation matters for the eventual contribution.
        \item \textit{Cross-paper relationship analysis}: How the works relate to each other, including whose ideas address whose limitations, what combinations open new possibilities, and how the logical chain across papers points toward an unexplored direction.
    \end{itemize}

    \item \textbf{Checkable Criteria} $\mathcal{U}_D = \{u_1, \ldots, u_K\}$: A set of verifiable items that a successful idea should satisfy, spanning three dimensions:
    \begin{itemize}[leftmargin=*,itemsep=1pt]
        \item \textit{Motivation criteria}: Does the output identify the key gap or limitation that $D$ addresses?
        \item \textit{Method criteria}: Does the proposed approach incorporate details from the prior works?
        \item \textit{Overall criteria}: Is the synthesis coherent, and does the proposed direction logically follow from the cross-paper analysis?
    \end{itemize}
\end{itemize}

\begin{figure}[t]
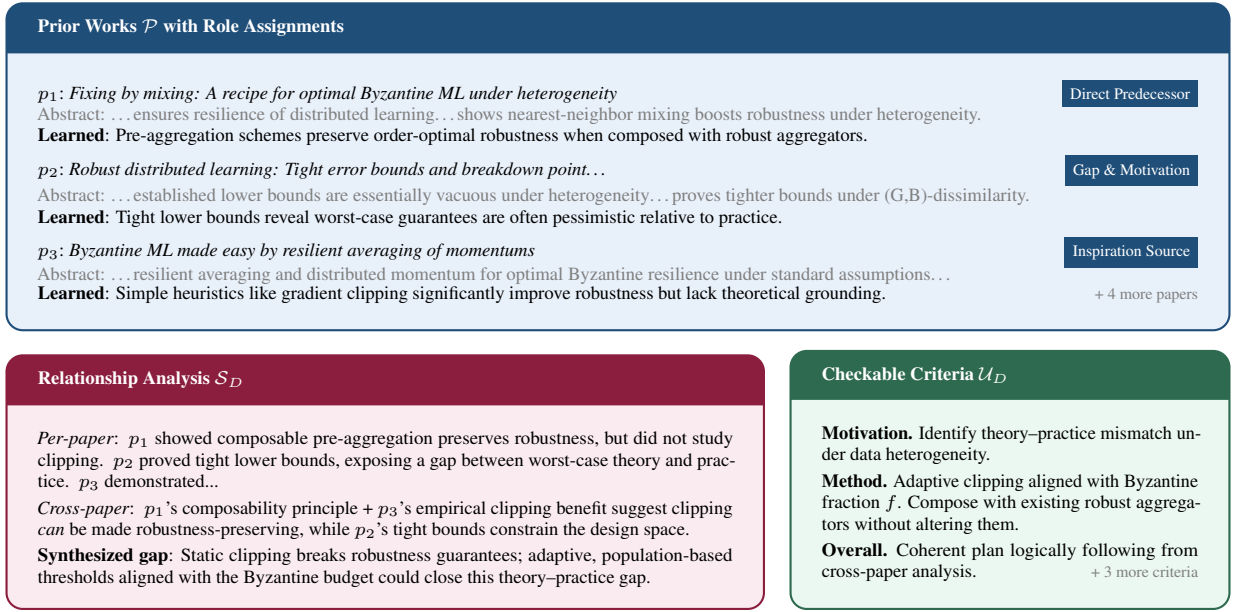

\centering
\scriptsize
\setlength{\parskip}{0pt}
% A figure cannot split across pages, so its boxes must be unbreakable.
\begin{headerbox}[Prior Works $\mathcal{P}$ with Role Assignments]{goalblue}{goalbluebg}[unbreakable]
$p_1$:~\textit{Fixing by mixing: A recipe for optimal Byzantine ML under heterogeneity} \hfill \colorbox{goalblue}{\textcolor{white}{\tiny Direct Predecessor}}\\
\textcolor{gray}{Abstract: \ldots ensures resilience of distributed learning\ldots shows nearest-neighbor mixing boosts robustness under heterogeneity.}\\
\textbf{Learned}: Pre-aggregation schemes preserve order-optimal robustness when composed with robust aggregators.\\[3pt]
$p_2$:~\textit{Robust distributed learning: Tight error bounds and breakdown point\ldots} \hfill \colorbox{goalblue}{\textcolor{white}{\tiny Gap \& Motivation}}\\
\textcolor{gray}{Abstract: \ldots established lower bounds are essentially vacuous under heterogeneity\ldots proves tighter bounds under (G,B)-dissimilarity.}\\
\textbf{Learned}: Tight lower bounds reveal worst-case guarantees are often pessimistic relative to practice.\\[3pt]
$p_3$:~\textit{Byzantine ML made easy by resilient averaging of momentums} \hfill \colorbox{goalblue}{\textcolor{white}{\tiny Inspiration Source}}\\[-1pt]
\textcolor{gray}{Abstract: \ldots resilient averaging and distributed momentum for optimal Byzantine resilience under standard assumptions\ldots}\\[-1pt]
\textbf{Learned}: Simple heuristics like gradient clipping significantly improve robustness but lack theoretical grounding.\hfill\textcolor{gray}{\tiny + 4 more papers}
\end{headerbox}

\vspace{1pt}

\noindent\begin{minipage}[t]{0.62\textwidth}
\begin{headerbox}[Relationship Analysis $\mathcal{S}_D$]{guidepurple}{guidepurplebg}[unbreakable]
\textit{Per-paper}:\enspace $p_1$ showed composable pre-aggregation preserves robustness, but did not study clipping.\enspace $p_2$ proved tight lower bounds, exposing a gap between worst-case theory and practice.\enspace $p_3$ demonstrated...\\[2pt]
\textit{Cross-paper}:\enspace $p_1$'s composability principle + $p_3$'s empirical clipping benefit suggest clipping \emph{can} be made robustness-preserving, while $p_2$'s tight bounds constrain the design space.\\[2pt]
\textbf{Synthesized gap}: Static clipping breaks robustness guarantees; adaptive, population-based thresholds aligned with the Byzantine budget could close this theory--practice gap.
\vspace{-2pt}
\end{headerbox}
\end{minipage}\hfill
\begin{minipage}[t]{0.36\textwidth}
\begin{headerbox}[Checkable Criteria $\mathcal{U}_D$]{rubricgreen}{rubricgreenbg}[unbreakable]
\textbf{Motivation.} Identify theory--practice mismatch under data heterogeneity.\\[2pt]
\textbf{Method.} Adaptive clipping aligned with Byzantine fraction $f$. Compose with existing robust aggregators without altering them.\\[2pt]
\textbf{Overall.} Coherent plan logically following from cross-paper analysis.
\hfill\textcolor{gray}{\tiny + 3 more criteria}
\end{headerbox}
\end{minipage}

\caption{A sample \anchor instance derived from a paper on Byzantine-resilient distributed learning. The anchor encodes per-paper role assignments with learned insights (\textit{top}), a cross-paper relationship analysis identifying the synthesized gap (\textit{bottom-left}), and literature-grounded checkable criteria (\textit{bottom-right}).}
\label{fig:idea_anchor}
\end{figure}

The value of \anchor is that they are \textbf{instance-specific and literature-grounded}: rather than generic quality criteria (\eg \textit{Is the idea novel?}), each anchor is derived from the particular way a real set of papers led to a real published contribution. This makes anchors a form of \textit{privileged information}~\citep{vapnik2009new}, knowledge available at training time (from the ground-truth paper) but not at inference, when the model must synthesize without knowing the answer. The anchor concept unifies the structured reasoning process, the rewards signal for RL, and the specification for high-quality demonstrations. By framing all three as aspects of a single underlying specification, we enable a comparison of different training strategies that exploit the same information source.

\subsection{Mining \anchor from Research Papers}
\label{sec:mining}

We construct \anchor at scale by reverse-engineering the intellectual genealogy of existing research papers. The pipeline is automated, using a sample creator model $\mathcal{M}_C$ to progressively build the structured specification from a ground-truth paper $D$ in three stages. \textbf{(i)~Prior work extraction.} Given the full text of $D$, $\mathcal{M}_C$ extracts the \textit{core idea} of $D$ and identifies 5--7 prior works that substantively shaped $D$'s contribution~\citep{beltagy2019scibert,lo2020s2orc}. Each prior work $p_i$ is assigned a functional role $\rho_i$ and a \textit{learned-insight} annotation as described in \autoref{sec:anchor_def}. \textbf{(ii)~Literature enrichment.} The extracted prior works are queried against scholarly search engines to retrieve their abstracts~\citep{priem2022openalex}, providing factual grounding beyond how $D$ describes them. \textbf{(iii)~Criteria generation.} Given the prior works, corresponding analysis, and reference proposal, $\mathcal{M}_C$ generates several checkable criteria $\mathcal{U}_D$ spanning motivation, method, and overall dimensions, ensuring that evaluation criteria are anchored in the specific intellectual genealogy of $D$. The full prompts for each stage are provided in~\autoref{app:prompts}.

\paragraph{Author Validation.} Beyond being topically related to $D$, the mined prior works are expected to be the ones that actually shaped its core idea. To verify this, we sent authors of benchmark papers the extracted prior works together with the reconstructed idea-formation path, and asked whether these reflect how their idea formed, which important works are missing, and which included works were not important. We received responses from the authors of 20 papers, 16 of whom were satisfied with the mined set. Five authors named one to three missing works (9 in total), so the mined sets cover 93.7\% of the works the authors credit; two authors flagged included works as unimportant, amounting to only 3 of the 134 extracted works (2.2\%). The mined sets therefore closely reflect the prior works that authors themselves credit for their ideas.

\paragraph{Paper Corpus.} We apply the mining pipeline to two domains. \textbf{(i)~Machine Learning.} We collect 7{,}494 accepted papers from ICLR, ICML, and NeurIPS, spanning from 2023 to 2025. \textbf{(ii)~Natural Science.} 6{,}689 papers published from 2023 to 2025 on Nature Communications, spanning 71 major scientific subject such as physics, chemistry, and neuroscience.

\paragraph{Evaluation Benchmark.} To assess generalization, we construct a temporally held-out evaluation set from 924 papers accepted at ICLR 2026, including all the 224 oral presentations. For each paper, we apply the same pipeline to generate candidate prior works and rubric criteria, which are then reviewed and refined by expert annotators. Annotators verify the correctness of extracted prior works, adjust role assignments where necessary, and edit criteria items to ensure they are unambiguous and faithfully reflect the paper's contribution. We describe the annotation protocol and inter-annotator agreement in detail in \autoref{app:annotation}.

%% file: section/3-method.tex
\section{Boosting Research Ideation with \anchor}
\label{sec:train}

We leverage \anchor in boosting research ideation. During training (\autoref{sec:training_paradigms}), three complementary paradigms exploit anchors as prompting context, privileged input, and reward signal. At inference (\autoref{sec:inference_enhancements}), two enhancements extend the model beyond abstract-only reasoning: retrieval-augmented generation with extended-text depth, and automated topic-driven literature discovery~\citep{izacard2021leveraging,lewis2020retrieval,nakano2021webgpt}.

\subsection{Anchor-Driven Training}
\label{sec:training_paradigms}

All three strategies treat $\mathcal{A}_D$ as \textit{privileged information}~\citep{vapnik2009new}, which is available during training but absent at deployment, and differ in how they convert it into learning signal.

\paragraph{Anchor-Guided Demonstration.}

Direct use of privileged information is to steer a strong teacher model toward higher-quality demonstrations~\citep{hinton2015distilling,hsieh2023distilling}. Without structured guidance, teacher outputs  might be fluent but often fail to deeply engage with the input literature~\citep{goel2025training}; providing a specification alongside $\mathcal{P}$ yields demonstrations that are more grounded.
For each instance $(\mathcal{P}, \mathcal{A}_D)$, we prompt an external LLM $\mathcal{M}_{\text{ext}}$ with both inputs: $y_{\text{demo}} = \mathcal{M}_{\text{ext}}(\mathcal{P}, \mathcal{A}_D)$. The target model $\pi_\theta$ is trained on the expert output with the anchor removed:
\begin{equation}
    \mathcal{L}_{\text{demo}} = -\mathbb{E}_{(\mathcal{P}, y_{\text{demo}})} \left[ \log \pi_\theta(y_{\text{demo}} \mid \mathcal{P}) \right].
\end{equation}
The anchor's influence is thus embedded in the demonstration: $\pi_\theta$ learns to satisfy anchor criteria without ever observing $\mathcal{A}_D$.

\paragraph{Anchor-Based Self-Distillation.}
Self-distillation removes the external-teacher requirement by exploiting the asymmetry between a model's output quality \textit{with} versus \textit{without} privileged context. The privileged-conditioned model serves as its own teacher while the unprivileged model is the student~\cite{zhao2026self}.
We sample $y_{\text{self}} = \pi_\theta(\mathcal{P}, \mathcal{A}_D)$ and train $\pi_\theta$ on its own anchor-conditioned outputs with $\mathcal{A}_D$ removed:
\begin{equation}
    \mathcal{L}_{\text{self}} = -\mathbb{E}_{(\mathcal{P}, y_{\text{self}})} \left[ \log \pi_\theta(y_{\text{self}} \mid \mathcal{P}) \right].
\end{equation}
We use system persona to prevent the model from revealing privileged information during thinking, avoiding training illusions~\citep{kim2026does}. We also compared different anti-leak strategies in~\autoref{app:leak}.

\paragraph{Anchor-Privileged Reinforcement Learning.}
Demonstration-based methods optimize a proxy (matching teacher output) rather than directly optimizing what defines a good idea. RL closes this gap: instance-specific privileged knowledge supplies a reward that is both more informative and harder to game than generic criteria.
A judge $\theta_r$ receives $(\mathcal{P}, y, \mathcal{A}_D)$ and scores each checkable item $u\in\mathcal{U}_D$:
\begin{equation}
    r_{\text{anchor}} = \frac{1}{|\mathcal{U}_D|} \sum_{u \in \mathcal{U}_D} \mathbb{I}[\theta_r(\mathcal{P}, y, u) = \text{satisfied}].
\end{equation}
Each $u_i$ is deemed satisfied only if no general quality guideline $\Gamma$ (specificity, soundness, feasibility) is violated. We optimize with GRPO~\citep{shao2024deepseekmathpushinglimitsmathematical}: for each $\mathcal{P}$ the policy samples $G$ candidates, scored by $\text{reward}(y) = r_{\text{anchor}}(y) - \lambda \cdot \mathbb{I}\{\text{format violation}\}$, and updates toward higher-scoring ones. Training is initialized from the base model, while the judge is flexible to be either frozen or evolving during training.

\subsection{Inference-Time Enhancements}
\label{sec:inference_enhancements}

\noindent
\begin{minipage}[t]{0.47\textwidth}
\vspace{0pt}
Training internalizes \textit{creative synthesis}, gap recognition and direction formulation. These capabilities let the model effectively conceptualize promising research directions, but lack specific details to translate ideas into actionable plans. Producing actionable proposals requires \textit{detail elaboration}, which we address with two inference-time mechanisms.
\end{minipage}
\hfill
\begin{minipage}[t]{0.5\textwidth}
\vspace{-10pt}
\small

\refstepcounter{algorithm}
\label{alg:inference}

\hrule
\vspace{2pt}
\textbf{Algorithm \thealgorithm: Ideation Pipeline}
\vspace{2pt}
\hrule
\vspace{2pt}

\begin{algorithmic}[1]
\Require Papers $\mathcal{P}$ or keywords $q$;\; $\texttt{Search}$, $\texttt{Retrieve}$,\; $\pi_\theta$
\If{input is $q$}
    \State $\mathcal{P} \leftarrow \texttt{Search}(q, N)$
\EndIf
\State $\mathcal{T} \leftarrow \pi_\theta.\texttt{think}(\mathcal{P})$ \Comment{role \& relationship analysis}
\For{$p_i \in \mathcal{P}$ \textbf{with} full-text access}
    \State $d_i \leftarrow \texttt{Retrieve}\bigl(p_i,\, \texttt{ExtractRole}(\mathcal{T}, p_i)\bigr)$
\EndFor
\State \Return $\pi_\theta.\texttt{generate}\bigl(\{(p_i, d_i)\},\, \mathcal{T}\bigr)$
\end{algorithmic}

\vspace{2pt}
\hrule
\end{minipage}

\paragraph{Role-Aware Retrieval Augmentation.}
When only abstracts are available, the model lacks methodological depth for tangible plans. We address this via \textit{role-aware retrieval}: $\pi_\theta$ first generates a thinking trace $\mathcal{T}$ that assigns each $p_i$ a functional role $\rho_i$. The system then fetches targeted full-text sections, including methods for \textit{Key Methodology}, results for \textit{Primary Baseline}, and problem statements for \textit{Gap \& Motivation}, and produces the final proposal from the augmented set:
\begin{equation}
    (\mathcal{M}, \mathcal{R}) = \pi_\theta \bigl(\{(p_i, d_i)\}_{i=1}^{N},\; \mathcal{T}\bigr),
\end{equation}
where $d_i = \texttt{Retrieve}(p_i, \rho_i)$ denotes role-guided passages.

\paragraph{Automated Topic-Driven Ideation.}
The model also accepts a high-level topic $q$ as an alternative to specific papers. It queries a search API to retrieve $N$ candidate papers, then applies the same synthesis pipeline~\citep{priem2022openalex,yao2022react}.
This pipeline can be optionally combined with full-text retrieval to generate a proposal in an end-to-end manner. \autoref{alg:inference} unifies the two inference modes.

%% file: section/4-exp.tex
\section{Experiments}
\label{sec:exp}

We evaluate whether \anchor-driven training improves literature-conditional research ideation beyond prompting alone. Our experiments ask three questions: (i) whether specialized training improves over the \texttt{Qwen3-8B} base model and approaches strong proprietary models; (ii) how retrieval-augmented inference complements training; and (iii) whether the learned synthesis behavior transfers across training scale, domains, and open-ended topic-driven ideation.

\subsection{Experimental Setup}

\paragraph{Models and Training Data.}
All trained policies are initialized from \texttt{Qwen3-8B} and \texttt{Qwen3.5-9B}~\citep{yang2025qwen3,qwen3.5}. We use the mined \anchor corpus described in~\autoref{sec:mining}, each training instance consists of 5 to 7 prior papers represented by titles and abstracts, together with the privileged \anchor extracted from the target paper. The three training paradigms in~\autoref{sec:training_paradigms} differ only in how this privileged information is converted into learning signal.
For anchor-guided demonstrations, we prompt \texttt{GPT-5.4-mini} to get the structured proposal for fine-tuning. For self-distillation, the trained model itself is prompted with the anchor and then trained on its own anchor-conditioned output under the unprivileged input. For RL, candidate proposals are scored item-by-item against the anchor criteria by \texttt{GPT-5.4-mini} judge, and the scores are used as rewards to optimize the policy. All model outputs follow the same structured format: $y = (\mathcal{T}, \mathcal{M}, \mathcal{R})$. Prompts and details in~\autoref{app:prompt}.

\paragraph{Retrieval-Augmented Inference.}
At inference time we evaluate both abstract-only generation and two retrieval variants. \texttt{RAG-Full} follows the method in \autoref{sec:inference_enhancements}: the model first assigns functional roles to input papers during thinking, then retrieves role-relevant passages and generates from the augmented context.
\texttt{RAG-Summary} uses the same retrieved passages but compresses each paper into a targeted summary by merging the abstract with the retrieved evidence. This second variant keeps input length and style close to the training distribution while injecting more task-directed methodological information.

\paragraph{Automated Evaluation.}
For each instance in our evaluation benchmark described in~\autoref{sec:mining}, \texttt{GPT-5.4} judges the generated proposal against the expert-refined checkable criteria. Each criterion is marked as satisfied only if the proposal puts forward a concrete, literature-grounded claim that meets the criterion while maintaining basic soundness and feasibility. We report the criteria satisfaction rate (CSR), averaged across all rubric items in the benchmark.

\paragraph{Human Evaluation.}
While automated rubrics support large-scale assessment, they cannot fully capture subjective nuances in research ideation. We conduct blind pairwise human evaluation to complement automatic metrics. We compare proposals from \texttt{Qwen3-8B} base and corresponding trained variants under identical input. Experts rank overall strength and assess literature grounding, synthetic novelty, methodological specificity, and feasibility. Presentation order is randomized, with ties allowed. Full annotation guidelines and subset statistics are provided in~\autoref{app:human_ranking}.

\subsection{Main Results}

\begin{wrapfigure}{r}{0.48\textwidth}
\setlength{\abovecaptionskip}{0pt}
\setlength{\belowcaptionskip}{5pt}

\centering
\vspace{-3mm}
\includegraphics[width=\linewidth]{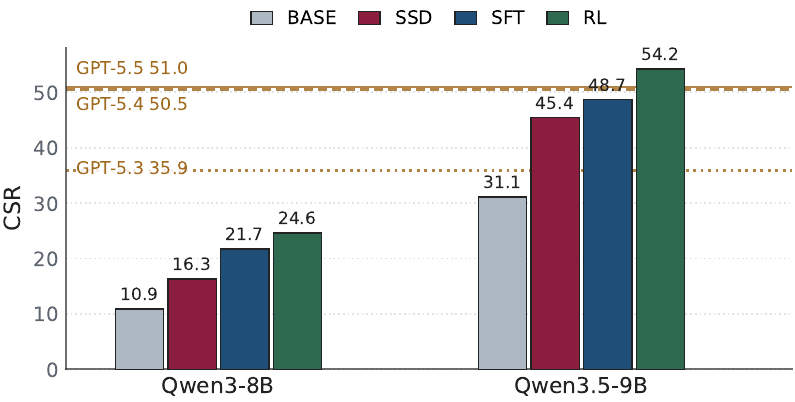}
\caption{Main automated evaluation on the ICLR 2026 benchmark with 924 instances. Bars report CSR (\%) for the base, self-distillation (SSD), SFT, and RL variants of \texttt{Qwen3-8B} and \texttt{Qwen3.5-9B}. Horizontal lines show proprietary model references.}
\vspace{-3mm}

\label{fig:main_results}
\end{wrapfigure}

\paragraph{Training with \anchor Improves Literature-Grounded Ideation.}
\autoref{fig:main_results} reports CSR on 924 ICLR instances, with proprietary models as upper references. Every training paradigm improves its base model. On \texttt{Qwen3-8B}, self-distillation reaches 16.3\% (+5.4), SFT 21.7\% (+10.8), and RL 24.6\% (+13.7), versus 10.9\% for the base. On \texttt{Qwen3.5-9B}, SFT and RL raise CSR from 31.1\% to 48.7\% and 54.2\%, respectively, with RL surpassing the references. Anchor supervision remains effective when the starting model is substantially stronger, rather than compensating for limited base-model capacity. RL gives the largest gain at both scales even though its reward contains only instance-specific criteria, suggesting that the policy learns a general pattern of gap identification and cross-paper synthesis instead of surface rubric matching. Retrieval adds smaller improvements, 25.6\% for RL with \texttt{RAG-Full} and 22.9\% for SFT with \texttt{RAG-Summary}, so access to more evidence cannot replace learned synthesis. We examine this difference in~\autoref{sec:analysis}.

\paragraph{Pairwise Ranking Confirms Holistic Quality Gains.}
Beyond item-wise rubric satisfaction, we conduct pairwise preference ranking on the same outputs. The judge receives two proposals, one from a trained variant and one from \texttt{Qwen3-8B} base, in randomized order and selects the one that presents better. Unlike CSR, which decomposes quality into independent checklist items, pairwise comparison captures holistic synthesis quality: whether a proposal constitutes a more coherent and actionable research direction given the same literature context.
\autoref{fig:ranking_results} reports win rates against the base under three judge configurations: LLM judge \texttt{GPT-5.4}, and human experts.
Despite variation in absolute magnitudes across paradigms and judges, the overall trend holds: all three training paradigms achieve $>$75\% pairwise win rate against the base model. RL leads, followed closely by SFT. Even self-distillation, which requires no external teacher, attains substantial preference gains, reinforcing that the anchor signal itself is the primary driver of improvement.

\subsection{Analysis and Case Studies}
\label{sec:analysis}

\input{figure_tex/exp_fig}

\paragraph{Retrieval Mostly Improves Methodological Detail.}
\begin{wraptable}{r}{0.42\textwidth}
\setlength{\abovecaptionskip}{0pt}
\setlength{\belowcaptionskip}{7pt}
\vspace{-24pt}
\centering
\small
\caption{Pairwise preference of RAG over abstract-only outputs. Each cell reports the win rate of the retrieval variant against the corresponding abstract-only output under a \texttt{GPT-5.4} judge.}
\label{tab:rag_pairwise}
\vspace{5mm}
\setlength{\tabcolsep}{6pt}
\begin{tabular}{lcc}
\toprule
\textbf{Variant} & \textbf{\texttt{RAG-Full}} & \textbf{\texttt{RAG-Sum.}} \\
\midrule
\texttt{SFT} & 73.6 & 74.2 \\
\texttt{SSD} & 68.7 & 70.3 \\
\texttt{RL}  & 76.8 & 75.5 \\
\bottomrule
\end{tabular}
\vspace{-1em}
\end{wraptable}

The two variants test an inference-time axis orthogonal to anchor training: \texttt{RAG-Full} exposes the model to role-relevant full text, whereas \texttt{RAG-Summary} compresses that evidence into an abstract-style input. \autoref{tab:rag_pairwise} shows that both are strongly preferred to abstract-only generation across all training paradigms, despite their modest CSR gains. We manually inspect 100 paired outputs to resolve this discrepancy. Retrieved passages add method choices, experimental settings, baselines, and implementation constraints, making proposals more actionable and improving holistic preference. They rarely change the high-level gap, motivation, or direction established earlier during \textit{creative synthesis}. Because CSR primarily rewards alignment with that direction, extra technical detail cannot recover a proposal whose synthesis is wrong.

This distinction also explains two automated trends. Retrieval slightly hurts the base model because unreliable role assignment can surface evidence for a weak premise and amplify it. Conversely, \texttt{RAG-Full} slightly outperforms the distribution-matched \texttt{RAG-Summary} for the strongest RL policy: once the model chooses a sound direction, richer full-text evidence becomes more useful than avoiding distribution shift. Retrieval is therefore best at elaborating a direction the model has already formulated, not deciding that direction from scratch.
Accordingly, CSR and pairwise preference measure complementary stages of the pipeline: the former emphasizes selecting the right research direction, while the latter also rewards how fully that direction is operationalized.

\paragraph{Cross-Domain Generalization and Scaling.}
\autoref{fig:training_dynamics} traces CSR throughout training for all three anchor-driven paradigms, with solid curves trained on machine-learning papers and dashed curves trained only on natural-science papers but evaluated on the same ICLR benchmark. In-domain training improves steadily with scale: SFT and RL exhibit large gains over the base model, while self-distillation yields a smaller but consistent increase, suggesting that additional anchor supervision strengthens literature-grounded synthesis rather than merely fitting the evaluation set. The models trained on natural-science \anchor also improve on the machine-learning ideation quality across all paradigms, despite never observing ML anchors during training. This transfer indicates that \anchor teach a domain-general ideation procedure: identifying a gap in prior work, relating evidence across papers, and formulating a motivated resolution, rather than memorizing domain-specific topics or surface patterns.

\paragraph{From-Scratch Ideation.}
Finally, we test whether the trained model can be used when no curated input papers are provided. We run a small topic-driven study that asks a narrower question: can the full pipeline move from an open research area to a literature-grounded proposal that experts find actionable?
We select six broad machine-learning topics, prompt the model to generate search queries for each topic, retrieve recent papers through Semantic Scholar API~\citep{kinney2023semantic}, and let the model select 6 to 8 papers per topic from the returned candidates based on topical relevance and diversity.
The \texttt{RL} policy with \texttt{RAG-Full} then reasoning over the retrieved papers, generates multiple candidate proposals, and uses a lightweight pairwise ranker to select the final output. Three ML researchers conduct blind review on a 1 to 5 scale for novelty, literature grounding, methodological specificity, feasibility, and overall quality.

\autoref{tab:from_scratch} suggests that anchor training remains useful beyond the benchmark setting. The full system improves most on literature grounding and methodological specificity, while maintaining feasibility. Qualitative reviews indicate that the main failure mode is not lack of retrieved evidence but weak selection among competing directions: lower-scoring proposals often combine individually relevant papers without committing to a sharp gap. In contrast, the stronger outputs first identify a non-obvious tension across the retrieved literature and then use full-text evidence to instantiate an experiment plan. This aligns with our benchmark analysis.

\begin{table}[H]
\centering
\small
\caption{Human evaluation of topic-driven ideation. Scores are averaged over proposals from six open ML topics.}
\label{tab:from_scratch}
\vspace{1mm}
\setlength{\tabcolsep}{4.5pt}
\begin{tabular}{lccccc}
\toprule
\textbf{Ideation System} & \textbf{Novelty} & \textbf{Grounding} & \textbf{Specificity} & \textbf{Feasibility} & \textbf{Overall} \\
\midrule
\texttt{Qwen3-8B} & 3.1 & 2.8 & 3.2 & 3.0 & 2.9 \\
\texttt{Qwen3-8B-RL w/ \texttt{RAG-Full}}   & 3.5 & 3.9 & 3.8 & 3.2 & 3.7 \\
\bottomrule
\end{tabular}
\end{table}

%% file: figure_tex/exp_fig.tex
\begin{figure}[t]
    \centering
    \begin{minipage}[t]{0.485\linewidth}
        \centering
        \vspace{0pt}
        \includegraphics[height=3.6cm]{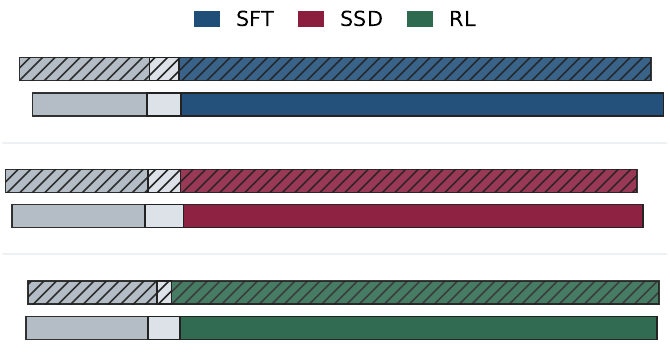}
        \caption{Pairwise preference over the base model.
        Each stacked bar compares a trained \texttt{Qwen3-8B} variant with the base model. Colored segments denote trained-model wins, gray segments denote base wins, and the center segment denotes ties. Hatched bars are \texttt{GPT-5.4} judgments and solid bars are human evaluations.}
        \label{fig:ranking_results}
    \end{minipage}
    \hfill
    \begin{minipage}[t]{0.485\linewidth}
        \centering
        \vspace{0pt}
        \includegraphics[height=3.6cm]{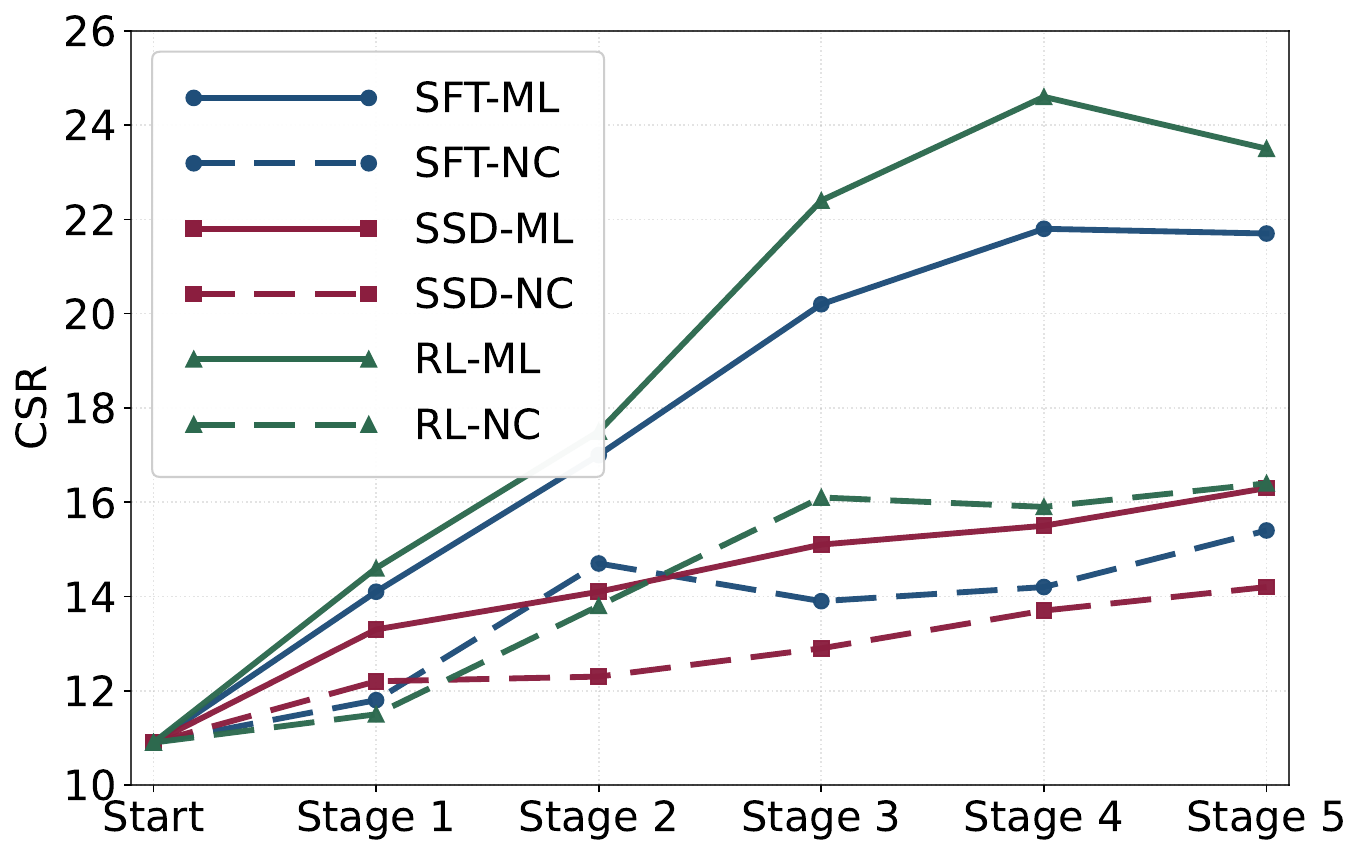}
        \caption{Training dynamics of anchor-driven ideation.
        CSR is measured on the ICLR benchmark across training stages for SFT, Self-Distillation, and RL over \texttt{Qwen3-8B}. Solid curves use machine-learning anchors, while dashed curves use only natural-science anchors and are evaluated on the same benchmark.}
        \label{fig:training_dynamics}
    \end{minipage}
\end{figure}

%% file: section/5-related.tex
\section{Related Work}

\paragraph{AI for Scientific Discovery.} LLMs are increasingly deployed across the scientific research pipeline~\citep{wang2023scientific}, assisting with literature search and synthesis~\citep{skarlinski2024language,zhao2025sciarena}, research code generation~\citep{wijk2024re,nathani2025mlgym}, and automated peer review~\citep{d2024marg,liang2024can}. Agents perform end-to-end experiment execution by optimizing objectives in sandboxed environments~\citep{boiko2023autonomous,jansen2025codescientist,novikov2025alphaevolvecodingagentscientific}, while multi-agent systems aim to automate workflows from hypothesis generation to paper writing~\citep{gottweis2025towards,schmidgall2025agentrxiv,schmidgall2025agent,yamada2025ai}. For idea generation, \citet{si2024llmsgeneratenovelresearch} provide the first large-scale human evaluation, finding LLM-generated ideas novel but weaker on feasibility. Follow-up systems refine ideas with reviewing agents over academic graphs~\citep{baek2024researchagent}, retrieve prior-paper inspirations and optimize novelty~\citep{wang2024scimon}, search diverse plans~\citep{hu2024nova}, evolve and compose idea facets~\citep{pu2025ideasynth}, recombine extracted facets with human-in-the-loop novelty verification~\citep{radensky2026scideator}, or invert single-concept assumptions via structured schemas~\citep{o2025sparks}. AInstein~\citep{mishra2025ainsteinassessingfeasibilityaigenerated} and follow-up studies~\citep{gupta2025all} further assess feasibility and quality. More recently, CHIMERA~\citep{sternlicht2026chimera} mines a large-scale knowledge base of pairwise concept recombinations from literature and trains a hypothesis generation model on it. However, most of these methods primarily rely on prompting strategies or single-paper manipulation at inference time, without explicit training for multi-paper synthesis.

\paragraph{Training with Privileged Information.} \anchor builds on the principle of learning using privileged information: additional information available only during training can guide learning~\citep{vapnik2009new}. In the context of LLMs, RL with LLM-graded rubrics extends training beyond verifiable domains~\citep{ouyang2022training}. \citet{goel2025training} use goal-specific rubrics as privileged information for self-grader to train research plan generators. LDC~\citep{li2024ldc} trains idea generators via SFT on paper-derived pairs and controllable RL with multi-dimensional reward models. Extensions include domain-specific rubric rewards~\citep{gunjal2025rubricsrewardsreinforcementlearning}, evolving rubrics that co-evolve with the policy~\citep{shao2025dr}, rubrics as dual-purpose exploration scaffolding and rewards~\citep{zhou2025breaking}, and rubric refinement via recursive decomposition~\citep{shen2026rethinking}. Recent work explores self-improvement where teacher and student share the same weights but differ in input access~\citep{zhao2026self}. $\pi$-Distill~\citep{penaloza2026privileged} jointly trains a privileged-conditioned teacher and an unconditioned student; GATES~\citep{stein2026gates} gates distillation on consensus among privileged rollouts; HDPO~\citep{ding2026hdpo} augments RL with privileged self-distillation on failures.

%% file: section/6-conclusion.tex
\section{Conclusion and Discussion}

We presented \anchor, a framework that mines structured, instance-specific specifications from published papers and uses them as privileged training signals for literature-grounded research ideation. Three complementary paradigms: demonstration, self-distillation, and reinforcement learning exploit these specifications, while role-aware retrieval augmentation enhances generation at inference time. Our analysis reveals a functional decomposition: creative synthesis benefits from training, whereas detail elaboration is driven by retrieval.

Our results also suggest several directions for future work. (i)~\textit{Selecting the research gap.} Anchor-based training improves ideation, whereas retrieval mainly adds methodological details without changing the gap a proposal targets. In the from-scratch study, low-scoring proposals combined relevant papers without committing to a clear gap. Training models to propose multiple candidate gaps and rank them is therefore a promising next step. (ii)~\textit{Multiple valid ideas.} Each instance uses one published paper as its target, but the same prior works might lead to several valid ideas. The anchor guides the model toward one well-grounded direction, but this does not mean that other directions are worse. Using several papers that build on the same prior works as targets, and measuring the diversity of generated ideas, would be promising~\citep{chen2026measuring,deng2026examining}. (iii)~\textit{Execution-level validation.} Our evaluation focuses on written proposals, while LLM ideas can lose more score than human ideas once they are executed~\citep{si2025ideation}. Implementing generated ideas and evaluating their experimental results is thus an important direction, although such validation is not feasible in every domain. 

\section*{Acknowledgments}
This work was supported in part by the U.S. National Science Foundation under award No. 2541654 and the Tata Consultancy Services.

%% file: section/X-suppl.tex
\section{Prompts for \anchor Mining}
\label{app:prompts}

This section provides the full prompts used in the \anchor mining pipeline described in~\autoref{sec:mining}. The pipeline consists of three prompt-driven stages: (\autoref{app:prompt_extraction})~extracting prior works with role assignments and learned insights from a ground-truth paper, (\autoref{app:prompt_synthesis})~generating a synthesis narrative and reference proposal that reconstructs the reasoning from prior works to the new idea, and (\autoref{app:prompt_criteria})~generating checkable evaluation criteria grounded in the prior works and reference proposal.

\subsection{Stage 1: Prior Work Extraction \& Role Assignment}
\label{app:prompt_extraction}

Given the full text of a ground-truth paper $D$, the creator model $\mathcal{M}_C$ extracts the core idea and identifies 5--7 prior works with functional role assignments and learned-insight annotations. The prompt enforces a \textbf{recency principle}: selected papers should be dominated by recent works (within ${\sim}3$ years of $D$), filtering out classic foundational citations that serve as background rather than proximal inspiration. Each candidate paper must pass three tests---counterfactual (would the idea exist without it?), specificity (what concrete element fed into $D$?), and proximity (is this the most recent version of the idea?).

\begin{headerbox}[Prompt: Prior Work Extraction \& Role Assignment]{neutralgray}{neutralgraybg}
\small
You are an expert AI research analyst specializing in tracing the \textbf{proximal intellectual lineage} of research ideas. Your task is to read a paper and reconstruct the \textbf{recent reading list} that led the authors to their core idea---identifying the specific recent papers they were actively engaging with when the idea took shape.

\medskip
\textbf{The Recency Principle.}\enspace
Research ideas are born from a researcher's \textbf{recent} reading, not from textbook knowledge. When a researcher writes a new paper, the creative process typically looks like this: (1)~they read a cluster of recent papers (typically published within the last 1--3 years); (2)~they notice gaps, opportunities, or combinable ideas across those recent papers; (3)~they synthesize a new contribution that builds on, extends, or recombines elements from that recent cluster.

Classic/foundational papers (e.g., the original Transformer, GANs, ResNet) are \textbf{background knowledge}---they are cited for completeness, but they are NOT the proximal triggers of the new idea. Your output should be dominated by recent papers---ideally the majority published within ${\sim}3$ years before the current paper.

\medskip
\textbf{Step 1: Determine the Paper's Publication Context.}\enspace
Identify (or estimate) the publication year of the current paper. Use this to calibrate your selection.

\medskip
\textbf{Step 2: Extract the Core Idea.}\enspace
Articulate: (a)~What is the paper's core innovation? (b)~What existing approach does it improve upon or depart from? (c)~What is the key insight that makes this work non-obvious?

\medskip
\textbf{Step 3: Trace the Proximal Intellectual Lineage.}\enspace
Think about which recent papers the authors were actively reading and engaging with. For each candidate prior work, apply \textbf{all three tests}:
\begin{enumerate}[leftmargin=1.5em,itemsep=1pt,topsep=2pt]
\item \textbf{Counterfactual test}: If the authors had NOT read this paper, would they still have come up with this specific idea? If yes, do not include it.
\item \textbf{Specificity test}: What SPECIFIC element (technique, insight, finding, limitation) from this prior work fed into the current paper? Must be concrete.
\item \textbf{Proximity test}: Is this the most recent/proximal version of this idea the authors engaged with? Always prefer the newer, more proximal paper.
\end{enumerate}

\medskip
\textbf{DO NOT INCLUDE:}
\begin{itemize}[leftmargin=1.5em,itemsep=0pt,topsep=2pt]
\item Classic foundational works that are common knowledge in the field.
\item Any paper whose main contribution has been fully absorbed into common practice.
\item Generic infrastructure/tools (PyTorch, CUDA, standard libraries).
\item Papers that share the domain but don't directly influence the core idea.
\item Papers only mentioned in passing or used only as evaluation baselines.
\item ``Grandparent'' papers---if the current paper builds on Paper A (2023) which builds on Paper B (2018), include A, not B.
\end{itemize}

\medskip
\textbf{DO INCLUDE:}
\begin{itemize}[leftmargin=1.5em,itemsep=0pt,topsep=2pt]
\item The most direct recent predecessor---the specific recent system/method this work directly extends or improves upon (typically within 1--2 years).
\item Recent papers whose LIMITATIONS or GAPS the current paper explicitly addresses.
\item Recent papers whose specific IDEAS or TECHNIQUES are directly extended, adapted, or combined.
\item Papers from potentially different subfields whose recent application of an idea was ``borrowed'' or adapted here.
\end{itemize}

\medskip
\textbf{Role Classifications} (assign ONE per paper):
\begin{enumerate}[leftmargin=1.5em,itemsep=1pt,topsep=2pt]
\item \textbf{Direct Predecessor}: The most closely related recent prior work---the specific system/method this paper directly extends or improves upon. Typically 1--2 papers.
\item \textbf{Inspiration Source}: A recent paper whose specific idea or technique (possibly from a different context) directly sparked the current paper's innovation.
\item \textbf{Gap \& Motivation}: A recent paper whose explicit limitations, failures, or unsolved challenges directly motivated this research.
\item \textbf{Methodological Ingredient}: A paper that introduced a specific technique or formulation that the current paper directly uses or adapts as a building block. Should be the most proximal source.
\end{enumerate}

\textit{Distribution guidance}: Aim for roughly 1--2 Direct Predecessors, 2--3 Inspiration Sources, 1--2 Gap \& Motivation papers, and 0--1 Methodological Ingredients.

\medskip
\textbf{cite\_id Convention.}\enspace
Each prior work must have a \texttt{cite\_id} in the format: \texttt{firstauthorlastname\allowbreak +year\allowbreak +keyword}, all lowercase, no spaces or punctuation (e.g., \texttt{chen2020simclr}, \texttt{radford2021clip}).

\medskip
\textbf{Final Recency Check.}\enspace
Before finalizing: (1)~Are the majority of selected papers from within ${\sim}3$ years? (2)~For each paper older than 5 years: is there truly no more recent paper that conveyed the same insight? (3)~Am I including any ``textbook knowledge'' papers? Remove them.

\medskip
\textbf{Output} (JSON): For each prior work (5--7 papers), provide: role, cite\_id, title, authors, year, and a one-sentence description of what the authors learned and how it informed the current work.
\end{headerbox}

\subsection{Stage 2: Synthesis Narrative \& Reference Proposal}
\label{app:prompt_synthesis}

Given the extracted prior works (with abstracts, roles, and learned insights), the core idea, and access to the original paper, $\mathcal{M}_C$ generates a synthesis narrative that reconstructs the reasoning from prior works to the new idea, and a structured research proposal (motivation + method). The synthesis narrative serves as the relationship analysis $\mathcal{S}_D$ component of the \anchor, and the reference proposal provides grounding for criteria generation in Stage~3.

\begin{headerbox}[Prompt: Synthesis Narrative \& Reference Proposal]{neutralgray}{neutralgraybg}
\small
You are an expert research analyst and proposal writer. Your task is to produce TWO things given a set of prior works (with their abstracts), the core idea of a paper, and the original paper content: (1)~a \textbf{synthesis narrative} that reconstructs the reasoning process from prior works to the new idea, and (2)~a \textbf{research proposal} that presents the motivation and method.

\medskip
\textbf{Part 1: Synthesis Narrative} (400--600 words)

Write this narrative in the first person, as the researcher who wrote this paper. You are reconstructing their reasoning process---how they analyzed these prior works, identified gaps and opportunities, and synthesized a new approach. Reference each prior work using its \texttt{@cite\_id}. Only mention papers that appear in the provided prior works list.

\textit{Critical rules}: This is a structured record of analytical reasoning---how specific observations about prior work led to specific design decisions. Prioritize clarity and depth of reasoning over narrative polish. Use precise, direct analytical language. Avoid dramatic or emotional expressions. The narrative consists of exactly \textbf{three paragraphs}, separated by blank lines. Do NOT include phase labels or headers.

\begin{enumerate}[leftmargin=1.5em,itemsep=2pt,topsep=2pt]
\item \textbf{Paragraph 1---Analysis of individual prior works} (${\sim}$100--150 words): Analyze each key prior work individually. For each paper, state: (a)~what it concretely achieved or proposed, (b)~what technical limitation, gap, or opportunity it exposes, and (c)~why that observation matters for the current research direction.

\item \textbf{Paragraph 2---Cross-paper synthesis} (${\sim}$200--250 words): This paragraph should contain NO new per-paper analysis---instead, focus entirely on how the observations from Paragraph~1 logically connect to form a new approach. Make the reasoning chain explicit: Which specific idea from one paper directly addresses a limitation identified in another? What new possibility emerges when techniques from different works are combined? Why had this particular combination not been explored before?

\item \textbf{Paragraph 3---Problem and approach summary} (${\sim}$200--250 words): Distill the reasoning into two clear points: (1)~the research problem---identifying the specific gaps or limitations revealed by prior work, analyzing their underlying causes, and defining the precise research questions; (2)~the proposed approach at a high level---what the core idea is, which insights from the above analysis it draws on, and why this design is expected to work.
\end{enumerate}

\medskip
\textbf{Part 2: Research Proposal}

The proposal has exactly two sections:

\textit{Section 1---Research Motivation \& Problem Definition} (2--3 paragraphs): Define the specific research problem being addressed. Identify the concrete gap or limitation in existing approaches, citing which prior works (\texttt{@cite\_id}) reveal this gap. Explain why this gap matters and what opportunity it creates.

\textit{Section 2---Core Idea \& Methodology}: An overview paragraph followed immediately by a numbered component list.
\begin{itemize}[leftmargin=1.5em,itemsep=0pt,topsep=2pt]
\item \textit{Overview paragraph}: A concise high-level description of the proposed approach. State the core idea in 2--3 sentences, then briefly explain why this formulation addresses the identified gap.
\item \textit{Numbered method components}: Break the method into 3--5 key technical components. Each component follows the format: \texttt{N.~\textbf{[Component Name]:} [Description of what this component does, why it is needed, and how it draws from or extends prior work. Reference relevant @cite\_id.]}
\end{itemize}

After the numbered list, add 1--2 sentences explaining why these components work together as a coherent system.

\medskip
\textbf{Writing guidelines}: Use \texttt{@cite\_id} throughout. Be technically precise. Write in a propositional tone (``We propose\ldots''). Keep the proposal to approximately 400--600 words.

\medskip
\textbf{Output} (JSON): \texttt{synthesis\_narrative}, \texttt{motivation}, \texttt{method}.
\end{headerbox}

\subsection{Stage 3: Checkable Criteria Generation}
\label{app:prompt_criteria}

Given the extracted prior works, synthesis narrative, and reference proposal, $\mathcal{M}_C$ generates checkable criteria $\mathcal{U}_D$ spanning motivation, method, and overall dimensions. The criteria are designed to admit multiple valid proposals---they check whether a proposal meaningfully engages with the gaps and opportunities revealed by the prior works, rather than whether it reproduces the specific method from the reference proposal.

\begin{headerbox}[Prompt: Checkable Criteria Generation]{neutralgray}{neutralgraybg}
\small
You are an expert in designing evaluation criteria for research proposals. Your task is to generate a set of \textbf{rubric items} that can be used to score the quality of a research proposal generated from a given set of prior works.

\medskip
\textbf{Context.}\enspace
You are given: (1)~prior works---a set of papers with their roles, cite\_ids, and what was learned from each; (2)~a synthesis narrative---a reasoning process showing how insights from the prior works connect; (3)~a reference proposal---a ground-truth proposal that serves as ONE valid example but NOT the only valid answer.

The rubrics will be used in reinforcement learning to score model-generated proposals. The model receives the same set of prior works (titles + abstracts) and must produce a research proposal with motivation and method.

\medskip
\textbf{Critical Design Principles:}
\begin{enumerate}[leftmargin=1.5em,itemsep=2pt,topsep=2pt]
\item \textbf{Ground rubrics in the prior works, not in the specific reference proposal.} Rubrics should check whether the proposal meaningfully engages with the gaps, limitations, and opportunities revealed by the prior works. They should NOT check whether the proposal reproduces the exact method from the reference proposal.

\item \textbf{Allow multiple valid solutions.} The same set of prior works can inspire different research problems and different methods. Your rubrics should admit multiple high-scoring proposals. A proposal with a completely different method than the reference can still score full marks if it is well-reasoned and grounded.

\item \textbf{Motivation rubrics can be moderately specific; method rubrics must be high-level.} Motivation rubrics may check that the proposal identifies specific gaps that the prior works clearly reveal. Method rubrics should check for general properties (e.g., ``addresses the encoding challenge'') rather than specific techniques (e.g., ``uses Fourier features'').

\item \textbf{Each rubric must be independently evaluable as pass/fail.} A reward model will check each rubric against the generated proposal. The rubric must be clear enough that pass/fail can be determined from reading the proposal alone.
\end{enumerate}

\medskip
\textbf{Rubric Categories:}

\textit{Motivation Rubrics} (2--3 items): Does the proposal identify the core limitation or gap that the ``Gap \& Motivation'' papers reveal? Does it explain why this gap matters? Does it articulate what opportunity the combination of prior works creates?

\textit{Method Rubrics} (2--3 items): Frame these as ``what the method should achieve or address'' rather than ``what specific technique it should use.'' Does the method address the identified gap in a principled way? Does it draw on or extend ideas from the provided prior works? Does it handle specific technical challenges that the prior works surface?

\textit{Overall Quality Rubrics} (1--2 items): Is the proposal internally consistent (motivation leads to method)? Does the proposal present a feasible and complete research plan?

\medskip
\textbf{Writing guidelines for individual rubrics:}
\begin{itemize}[leftmargin=1.5em,itemsep=0pt,topsep=2pt]
\item Each criterion is ONE sentence. Be precise but not overlong.
\item For motivation rubrics: reference which prior work(s) reveal the gap using \texttt{@cite\_id}.
\item For method rubrics: describe what the method should \textit{achieve} or \textit{address}, not what specific technique it should use.
\item Avoid rubrics that can only be satisfied by one specific approach.
\item Make sure the full set of rubrics, taken together, does NOT uniquely determine a single method.
\end{itemize}

\medskip
\textbf{Output} (JSON): Generate 5--7 rubric items total, each with: \texttt{id}, \texttt{category} (motivation / method / overall), and \texttt{criterion}.
\end{headerbox}

\section{Prompts for Training Data Synthesis}
\label{app:prompt}

This section provides the full prompts used in the three anchor-driven training paradigms described in~\autoref{sec:training_paradigms}.

\subsection{Demonstration Generation Prompt (Shared by SFT and Self-Distillation)}
\label{app:prompt_demo}
\label{app:prompt_ssd}

Anchor-guided demonstration (SFT) and anchor-based self-distillation (SSD) share the same generation prompt: both paradigms condition on the input papers $\mathcal{P}$ together with the \anchor $\mathcal{A}_D$ and produce a structured proposal comprising a thinking trace, research motivation, and research plan. The two paradigms differ only in \textit{which model} executes the prompt---SFT uses the external teacher $\mathcal{M}_{\text{ext}}$ (\texttt{GPT-5.4-mini}), while SSD uses the target model $\pi_\theta$ (\texttt{Qwen3-8B}) itself---and in the anti-leakage system persona prepended for SSD (see~\autoref{app:leak}).

\begin{headerbox}[Prompt: Anchor-Conditioned Demonstration Generation]{neutralgray}{neutralgraybg}
\small

\textbf{System Message:}

\medskip

You are a research scientist skilled at synthesizing ideas from existing literature into novel research proposals. You are given a set of related research papers (with titles and abstracts). Your task is to analyze these papers, identify research gaps and opportunities, and propose a novel research idea. When referencing prior works, use its \texttt{@cite\_id} (e.g., \texttt{@chen2020simclr}).

\medskip

\textit{[Anchor-derived context inserted here --- presented directly as rubric criteria for SFT; dissolved into persona knowledge for SSD (see below)]}

\medskip

Analyze the papers below and write a research proposal with your own original thinking. Structure your response as:
\begin{enumerate}[leftmargin=2em]
\item \texttt{<motivation>} and \texttt{</motivation>} tags --- explain the research gap
\item \texttt{<method>} and \texttt{</method>} tags --- describe your proposed approach
\end{enumerate}

Place your actual content directly inside each tag pair.

\bigskip
\textbf{User Message:}

\medskip

\texttt{\# \{Paper Title 1\} (@cite\_id\_1)}\\
\texttt{\#\# Summary: \{Abstract 1\}}

\medskip

\texttt{\# \{Paper Title 2\} (@cite\_id\_2)}\\
\texttt{\#\# Summary: \{Abstract 2\}}

\medskip

\texttt{...}

\end{headerbox}

\vspace{4pt}

For self-distillation, the following system persona is prepended to prevent the model from leaking privileged anchor content into its thinking trace or proposal:

\begin{headerbox}[System Persona: Anti-Leakage Constraint (SSD only)]{neutralgray}{neutralgraybg}
\small

The anchor criteria $\mathcal{A}_D$ are not presented as explicit rubrics to the model. Instead, each criterion is rewritten as a first-person observation that the model has ``already made'' from reading the literature, avoiding all meta-language (e.g., ``criteria'', ``guidance'', ``rubric''). The dissolved form is injected into the system prompt as follows:

\bigskip

From your reading of the recent literature, you have observed several important trends and gaps: \textit{\{motivation insight 1\}}. Furthermore, \textit{\{motivation insight 2\}}. Furthermore, \textit{\{motivation insight 3\}}.

\medskip

Based on these observations, you believe a promising research direction would involve a method that would \textit{\{method insight 1\}}; \textit{\{method insight 2\}}; \textit{\{method insight 3\}}.

\bigskip

Each motivation-category rubric is stripped of prescriptive prefixes (e.g., ``The proposal should~\ldots'') and verb phrases (e.g., ``identify that~\ldots''), lowercased, and concatenated with ``Furthermore,~\ldots'' connectors. Each method-category rubric is similarly normalized and joined with semicolons. This prevents the model from reciting privileged anchor content verbatim during chain-of-thought reasoning.

\end{headerbox}

\subsection{RL Reward Judging Prompt}
\label{app:prompt_judge}

For anchor-privileged RL, the judge model receives a generated proposal alongside the anchor criteria and scores each checkable item as satisfied or not.

\begin{headerbox}[Prompt: RL Reward Judge (Per-Criterion Scoring)]{neutralgray}{neutralgraybg}
\small

\textbf{System Message:}

\medskip

You are a strict but fair evaluator of research proposals. You need to evaluate if the Proposed Research Plan satisfies the provided evaluation rubrics.

\medskip

\textbf{Rules}
\begin{enumerate}[leftmargin=2em]
\item Evaluate each rubric independently.
\item Be strict but reasonable --- semantic equivalence counts.
\item Focus on substance, not style.
\item Only score based on what is explicitly stated.
\item Be skeptical, careful, and come up with valid criticisms. Be as strict as possible, while being unbiased and reasonable.
\item Note that the plan should not just say it satisfies these desiderata, don't be fooled by that. Check carefully WHETHER, HOW and WHY the proposed plan meets each desiderata for this rubric item one by one.
\end{enumerate}

\medskip

\textbf{Output Format}\\
Return ONLY a JSON object. For each rubric, give a hit (1) or miss (0) and a brief reason. Then give the total.

\medskip

Example with 3 rubrics:
\begin{lstlisting}[language={},basicstyle=\ttfamily\footnotesize,frame=none]
{
  "rubrics": [
    {"id": 1, "reason": "Proposal clearly identifies X.", "hit": 1},
    {"id": 2, "reason": "No mention of Y.", "hit": 0},
    {"id": 3, "reason": "Method addresses Z.", "hit": 1}
  ],
  "total_score": 2,
  "max_score": 3
}
\end{lstlisting}

Do NOT output anything outside the JSON object. No markdown, no explanation before or after.

\bigskip
\textbf{User Message:}

\medskip

\textbf{Proposal}

\medskip

\textit{\{generated proposal text\}}

\medskip

\textbf{Rubrics}

\begin{itemize}[leftmargin=2em]
\item Rubric \textit{id} (\textit{category}): \textit{criterion text}
\item Rubric \textit{id} (\textit{category}): \textit{criterion text}
\item \ldots
\end{itemize}

Evaluate each rubric and return the JSON.

\end{headerbox}

\section{Prompts for Evaluation}
\label{app:prompt_eval}

\subsection{Criteria-Based Evaluation Prompt}
\label{app:prompt_eval_criteria}

For automated evaluation, \texttt{GPT-5.4} judges whether each criterion in the expert-refined checkable criteria $\mathcal{U}_D$ is satisfied by the generated proposal. Each criterion is marked as satisfied only if the proposal presents a concrete, literature-grounded claim that meets the criterion while maintaining soundness and feasibility.

\begin{headerbox}[Prompt: Criteria-Based Evaluation]{neutralgray}{neutralgraybg}
\small
You are a rigorous expert peer reviewer evaluating research proposals against specific rubrics. Your evaluation must reward depth of understanding over breadth of keyword coverage.

\vspace{0.3em}
\textbf{Core Principle} \\
A rubric should be marked HIT only when the proposal demonstrates \textbf{genuine technical understanding} of the criterion --- not merely surface-level mention of relevant terms. The most common failure mode is a proposal that lists many related techniques or concepts without clearly reasoning about them.

\vspace{0.3em}
\textbf{Rules}
\begin{enumerate}[leftmargin=1.5em, itemsep=0pt, topsep=2pt]
\item Evaluate each rubric independently.
\item HIT (1): The proposal \textbf{clearly and specifically} addresses the criterion with concrete technical substance. The proposal must show it understands WHY the criterion matters and HOW it is addressed --- not just THAT relevant terms appear.
\item MISS (0): The proposal fails to address the criterion, addresses it only superficially, or mentions relevant keywords without demonstrating understanding of the underlying technical point.
\item \textbf{Kitchen-sink penalty}: If a proposal lists many techniques, frameworks, or citations without clearly explaining each one's role and connection to the criterion, this is a sign of superficial coverage. Do NOT reward scattershot inclusion of buzzwords or methods. A proposal that names 10 techniques but explains none in depth should score lower than one that deeply explains 3 relevant techniques.
\item \textbf{Depth reward}: If a proposal addresses the criterion through a different but clearly valid technical approach with sound reasoning, mark as HIT even if terminology differs from the rubric. Genuine insight expressed in different words is more valuable than exact keyword matching without understanding.
\item For motivation rubrics: the proposal must show it understands the specific gap or opportunity --- not just mention the topic area. A well-reasoned argument identifying the core issue is worth more than a laundry list of related problems. If the proposal clearly explains WHY a gap exists and what its implications are, that demonstrates understanding even if framed differently.
\item For method rubrics: the proposal must describe a concrete, well-justified mechanism. A generic pipeline that happens to list a relevant technique does NOT satisfy the criterion --- the proposal must explain how that technique specifically serves the rubric's requirement. Conversely, a focused method that achieves the rubric's goal through a different but well-explained approach DOES satisfy it.
\item For overall/coherence rubrics: a focused, well-integrated plan where each component is clearly motivated is superior to an ambitious plan with many loosely connected components. Penalize proposals that stack borrowed techniques without explaining integration. Reward proposals where the stated problem clearly drives the method design, even if the method is simpler.
\item When in doubt, mark as MISS.
\end{enumerate}

\vspace{0.3em}
\textbf{Output Format} \\
Return ONLY a JSON object. For each rubric, give a hit (1) or miss (0) and a brief reason. Then give the total.

\vspace{0.3em}
\begin{lstlisting}[language={},basicstyle=\ttfamily\footnotesize,frame=none]
{
  "rubrics": [
    {"id": 1, "reason": "Proposal clearly identifies X.", "hit": 1},
    {"id": 2, "reason": "No mention of Y.", "hit": 0},
    {"id": 3, "reason": "Method addresses Z.", "hit": 1}
  ],
  "total_score": 2,
  "max_score": 3
}
\end{lstlisting}

\vspace{0.2em}
Do NOT output anything outside the JSON object. No markdown, no explanation before or after.
\end{headerbox}

\section{Annotation Protocol and Inter-Annotator Agreement}
\label{app:annotation}

The evaluation benchmark consists of 924 papers accepted at ICLR 2026, including all 224 oral presentations. For each paper, our mining pipeline generates candidate prior works, role assignments, and checkable criteria, which are then reviewed and refined by expert annotators.

\paragraph{Annotator Recruitment.}
We recruit annotators who are graduate students or researchers in machine learning with at least two years of research experience. Each annotator has published at least one first-author paper at a top-tier ML venue.

\paragraph{Annotation Procedure.}
Each instance is independently reviewed by two annotators. Annotators perform three tasks: (i)~verify the correctness of extracted prior works and remove irrelevant ones, (ii)~adjust role assignments where the automated pipeline misclassifies a paper's functional role, and (iii)~edit criteria items to ensure they are unambiguous, non-trivial, and faithfully reflect the paper's contribution. Disagreements are resolved through discussion, with a senior researcher adjudicating unresolved cases.

\section{Experimental Configuration}
\label{app:exp_config}

\paragraph{Hardware.}
Experiments are conducted on a 4 $\times$ NVIDIA B200 GPUs node.

\paragraph{SFT / Self-Distillation Hyperparameters.}
Since anchor-guided demonstration (SFT) and anchor-based self-distillation (SSD) differ only in the source model that generates the training data (see~\autoref{app:prompt_demo}), they share the same fine-tuning configuration. \autoref{tab:sft_config} summarizes the shared hyperparameters.

\begin{table}[h]
\centering
\small

\begin{minipage}[t]{0.48\textwidth}
\centering
\caption{\textbf{SFT / Self-distillation training configuration.}}
\label{tab:sft_config}
\vspace{1mm}
\begin{tabular}{ll}
\toprule
\textbf{Hyperparameter} & \textbf{Value} \\
\midrule
Base model & \texttt{Qwen3-8B} \\
Learning rate & $1 \times 10^{-5}$ \\
LR scheduler & Cosine \\
Batch size & 32 \\
Number of epochs & 5 \\
Max sequence length & 8192 \\
Warmup ratio & 0.1 \\
Weight decay & 0.1 \\
Precision & \texttt{bfloat16} \\
\midrule
\multicolumn{2}{l}{\textit{SFT-specific: demonstration source}} \\
Teacher model & \texttt{GPT-5.4-mini} \\
\midrule
\multicolumn{2}{l}{\textit{SSD-specific: demonstration source}} \\
Teacher model & \texttt{Qwen3-8B} \\
Anti-leakage & System persona constraint \\
\bottomrule
\end{tabular}
\end{minipage}
\hfill
\begin{minipage}[t]{0.48\textwidth}
\centering
\caption{\textbf{RL training configuration.}}
\label{tab:rl_config}
\vspace{1mm}
\begin{tabular}{ll}
\toprule
\textbf{Hyperparameter} & \textbf{Value} \\
\midrule
Base model & \texttt{Qwen3-8B} \\
Learning rate & $1 \times 10^{-6}$ \\
LR scheduler & Cosine \\
Batch size & 512 \\
Group size $G$ & 8 \\
Number of training steps & 150 \\
Max sequence length & 8192 / 2048 \\
KL coefficient $\beta$ & 0.0 \\
Clip $\epsilon$ & 0.2 \\
Format penalty $\lambda$ & $-1.0$ \\
Reward model & \texttt{GPT-5.4-mini} \\
Precision & \texttt{bfloat16} \\
\bottomrule
\end{tabular}
\end{minipage}

\end{table}

\paragraph{Inference Configuration.}
During RL rollout, completions are sampled with temperature $0.7$, top-$k$ = 50, top-$p$ = 0.9, and a maximum generation length of 2048 tokens. At evaluation time, greedy decoding is used (temperature $0.0$) with a maximum of 64 new tokens for SFT/SSD models.

\section{Training and Evaluation Data Statistics}
\label{app:data_stats}

\paragraph{Training Corpus Statistics.}

\autoref{tab:train_stats} summarizes the training corpus across the two domains. Each instance contains 5--8 input prior works and 6--7 checkable criteria generated by the mining pipeline. The total corpus contains 14{,}183 instances with 96{,}667 criteria.

\begin{table}[h]
\centering
\small
\caption{\textbf{Training corpus statistics.}}
\label{tab:train_stats}
\vspace{1mm}
\setlength{\tabcolsep}{4pt}
\resizebox{\textwidth}{!}{
\begin{tabular}{lccccc}
\toprule
\textbf{Domain} & \textbf{\# Papers} & \textbf{Avg. Prior Works} & \textbf{Avg. Criteria} & \textbf{Total Criteria} & \textbf{Source Venues} \\
\midrule
Machine Learning & 7{,}494 & 6.52 & 6.86 & 51{,}416 & ICLR, ICML, NeurIPS (2023--2025) \\
Natural Science  & 6{,}689 & 5.90 & 6.76 & 45{,}251 & Nature Comm. (2023--2025) \\
\midrule
Total            & 14{,}183 & 6.23 & 6.82 & 96{,}667 & --- \\
\bottomrule
\end{tabular}
}
\end{table}

\paragraph{Evaluation Benchmark Statistics.}

\autoref{tab:eval_stats} provides detailed statistics for the ICLR 2026 evaluation benchmark. Each instance is annotated with 7--8 expert-refined criteria across three dimensions.

\begin{table}[h]
\centering
\small
\caption{\textbf{ICLR 2026 evaluation benchmark statistics.}}
\label{tab:eval_stats}
\vspace{1mm}
\begin{tabular}{lclclc}
\toprule
\multicolumn{2}{c}{\textbf{Benchmark size}} &
\multicolumn{2}{c}{\textbf{Criteria}} &
\multicolumn{2}{c}{\textbf{Criteria type}} \\
\cmidrule(lr){1-2}
\cmidrule(lr){3-4}
\cmidrule(lr){5-6}
\textbf{Statistic} & \textbf{Value} &
\textbf{Statistic} & \textbf{Value} &
\textbf{Statistic} & \textbf{Value} \\
\midrule
Total instances & 924
& Total checkable criteria & 6{,}667
& Avg. motivation criteria & 2.99 \\

Oral presentations & 224
& Avg. criteria per instance & 7.22
& Avg. method criteria & 3.22 \\

Spotlight / poster & 700
& Instances with 7 criteria & 725
& Avg. overall criteria & 1.00 \\

Total input papers & 6{,}108
& Instances with 8 criteria & 199
&  &  \\

Avg. prior works per instance & 6.61
&  & 
&  &  \\

Range & 4--8
&  &
&  &  \\
\bottomrule
\end{tabular}
\end{table}

\paragraph{Criteria Distribution across Training Domains.}

\autoref{tab:criteria_dist} compares the distribution of criteria categories across the two training domains. Both domains show a balanced distribution across motivation and method criteria, with exactly one overall criterion per instance on average.

\begin{table}[h]
\centering
\small
\caption{\textbf{Per-category criteria distribution across training domains.}}
\label{tab:criteria_dist}
\vspace{1mm}
\begin{tabular}{lccc}
\toprule
\textbf{Domain} & \textbf{Avg. Motivation} & \textbf{Avg. Method} & \textbf{Avg. Overall} \\
\midrule
Machine Learning & 2.97 & 2.89 & 1.01 \\
Natural Science  & 2.71 & 3.00 & 1.06 \\
Evaluation (ICLR 2026) & 2.99 & 3.22 & 1.00 \\
\bottomrule
\end{tabular}
\end{table}

\paragraph{Criteria Count Distribution across Training Domains.}

\begin{table}[h]
\centering
\small
\caption{\textbf{Criteria count distribution.} Number of instances with each rubric count.}
\label{tab:rubric_dist}
\vspace{1mm}
\begin{tabular}{lccc}
\toprule
\textbf{Criteria count} & \textbf{ML} & \textbf{Natural Science} & \textbf{Evaluation} \\
\midrule
6 criteria & 1{,}042 & 1{,}572 & 0 \\
7 criteria & 6{,}452 & 5{,}117 & 725 \\
8 criteria & 0 & 0 & 199 \\
\bottomrule
\end{tabular}
\end{table}

\section{Human Evaluation Details}
\label{app:human_ranking}

\paragraph{Evaluation Protocol.}
We conduct blind pairwise human evaluation to complement automatic metrics. For each evaluation instance, annotators receive the same set of input papers $\mathcal{P}$ and two proposals generated by different model variants (e.g., \texttt{Qwen3-8B Base} vs.\ a trained variant). The presentation order of the two proposals is randomized to eliminate position bias.

\paragraph{Evaluation Dimensions.}
Annotators assess proposals along five dimensions:
\begin{itemize}[leftmargin=*, itemsep=2pt]
    \item \textbf{Overall quality}: Which proposal presents a stronger, more promising research direction?
    \item \textbf{Literature grounding}: Which proposal better leverages the input papers and demonstrates deeper understanding of the prior work?
    \item \textbf{Synthetic novelty}: Which proposal identifies a more creative and non-obvious gap or research direction from the combination of input papers?
    \item \textbf{Methodological specificity}: Which proposal provides a more concrete and actionable research plan with specific technical details?
    \item \textbf{Feasibility}: Which proposal outlines a more realistic and executable research agenda?
\end{itemize}

\noindent For each dimension, annotators select one of three options: \textit{Proposal A is better}, \textit{Proposal B is better}, or \textit{Tie}. Annotators are encouraged to provide brief justifications for their choices.

\section{Self-Distillation Data Leakage Analysis}
\label{app:leak}

In anchor-based self-distillation (\autoref{sec:training_paradigms}), the model is prompted with the \anchor as privileged context during data generation. A key risk is that the model may ``leak'' privileged information---directly copying anchor content into its thinking trace or proposal---rather than internalizing the synthesis pattern. If such leakage occurs, the resulting training data would contain references to ground-truth answers that are unavailable at inference time, creating a training--inference mismatch.

\paragraph{Anti-Leak Strategies.}
We compare six strategies that vary in how the anchor is presented and how the model is instructed to use it, against a naive baseline with no anti-leak protection:

\begin{itemize}[leftmargin=*, itemsep=2pt]
    \item \textbf{Naive (baseline)}: The anchor is placed directly in the prompt with no structural tags or anti-leak instructions. The model generates free-form output.
    \item \textbf{Persona knowledge}: The anchor is reframed as the model's internalized domain expertise rather than an external reference, discouraging verbatim copying while preserving full semantic access.
    \item \textbf{Bottleneck PI}: The anchor is compressed into single-keyword dimensions, maximally restricting verbatim content while transmitting abstract directional signals.
    \item \textbf{Quality characteristics}: The anchor is dissolved into a set of desired output quality attributes, preserving directional information while removing copyable details.
    \item \textbf{Critique--edit}: A two-phase pipeline where the model first generates independently, then self-critiques against the anchor within a \texttt{<think>} block and revises.
    \item \textbf{Uncertainty steering}: A three-stage thinking structure (explore $\to$ reflect $\to$ refine) that channels the anchor through structured uncertainty rather than direct reference.
    \item \textbf{Weighted SFT}: Only emphasis weights over rubric dimensions are transmitted, without explicit content signal from the anchor.
\end{itemize}

\paragraph{Leakage Detection Metrics.}
We measure leakage using: (i)~keyword pattern matching for telltale phrases that reference the anchor (e.g., ``the rubric says,'' ``according to the criteria''), (ii)~structural tag validation (whether the output correctly uses \texttt{<motivation>}/\texttt{<method>} tags), and (iii)~manual inspection of all 50 samples per strategy. An output is marked as \emph{leaked} if it directly references or quotes anchor content that would be unavailable at inference time.

\paragraph{Results.}
We evaluate all strategies on 50 samples (10 shared problems $\times$ 5 generations each) using Qwen3-8B with vLLM inference. We measure three complementary metrics: leakage rate, rubric compliance (via LLM-as-judge), and ROUGE-L against the ground-truth reference.

\begin{table}[h]
\centering
\small
\caption{\textbf{Comparison of anti-leak strategies for self-distillation.} Leakage = fraction of outputs that reference anchor content; Rubric Compliance = LLM-as-judge pass rate over motivation/method/overall criteria; ROUGE-L = overlap with ground-truth reference. $^\dagger$False positives: the word ``criterion'' used in normal academic context.}
\label{tab:leak_comparison}
\vspace{1mm}
\begin{tabular}{lcccc}
\toprule
\textbf{Strategy} & \textbf{Leakage $\downarrow$} & \textbf{Rubric Compl.\ $\uparrow$} & \textbf{ROUGE-L $\uparrow$} & \textbf{Rank} \\
\midrule
Naive (baseline)            & 100\% (50/50) & 0.836 & 0.147 & 5 \\
\midrule
Persona knowledge           & \textbf{0\%} (0/50)    & \textbf{1.000} & \textbf{0.193} & \textbf{2} \\
Bottleneck PI               & \textbf{0\%} (0/50)    & 0.522 & 0.164 & 7 \\
Quality characteristics     & 4\% (2/50)    & \textbf{1.000} & 0.192 & \textbf{1} \\
Critique--edit              & 4\%$^\dagger$ (2/50)   & \textbf{1.000} & 0.193 & \textbf{3} \\
Uncertainty steering        & 4\%$^\dagger$ (2/50)   & 0.955 & 0.186 & 4 \\
Weighted SFT                & 2\% (1/50)    & 0.701 & 0.156 & 6 \\
\bottomrule
\end{tabular}
\end{table}

\paragraph{Discussion.}

Three findings emerge from \autoref{tab:leak_comparison}:

\begin{itemize}[leftmargin=*, itemsep=2pt]
    \item \textbf{No anti-leak protection yields complete leakage.} The naive baseline leaks in 100\% of outputs. Without structural tags or instructions, the model freely copies and references anchor content, producing outputs that cannot be used as training data due to training--inference mismatch.

    \item \textbf{Zero leakage is achievable without sacrificing quality---but not always.} Persona knowledge achieves 0\% leakage while maintaining perfect rubric compliance (1.000) and the highest ROUGE-L (0.193), demonstrating the ideal safety--quality balance. In contrast, bottleneck PI also achieves 0\% leakage but at severe cost to rubric compliance (0.522), because compressing the anchor into single-keyword dimensions strips too much semantic content for the model to produce adequate outputs.

    \item \textbf{Remaining leakage is largely spurious.} For critique--edit and uncertainty steering, the detected 4\% leakage (2/50 each) consists of false positives---the word ``criterion'' used in normal academic prose, not as a reference to anchor rubrics. After accounting for false positives, five of six strategies achieve effectively 0--2\% true leakage, confirming that structured prompting reliably prevents information copying.
\end{itemize}

Among strategies with near-zero leakage, the quality--leakage frontier is dominated by three approaches: \textbf{quality\_chars} (best ROUGE-1 at 0.543, perfect compliance), \textbf{persona\_knowledge} (best safety profile with 0\% leakage and 0 tag errors), and \textbf{critique\_edit} (perfect compliance with self-contained reasoning in \texttt{<think>}). The choice among them depends on whether the practitioner prioritizes absolute zero leakage (persona knowledge), maximal ground-truth similarity (quality characteristics), or interpretable self-critique traces (critique--edit).

\section{Additional Judge Evaluation}
\label{app:more_judge}

To verify that our automated evaluation results are not dependent on a specific LLM judge, we repeat the criteria satisfaction rate (CSR) evaluation with alternative judge models.

\begin{table}[h]
\centering
\small
\caption{\textbf{CSR evaluated by different LLM judges.} All results on the 924-instance ICLR 2026 benchmark (abstract-only setting).}
\label{tab:judge_comparison}
\vspace{1mm}
\begin{tabular}{lcccc}
\toprule
\textbf{Judge} & \textbf{Base} & \textbf{SFT} & \textbf{SSD} & \textbf{RL} \\
\midrule
\texttt{GPT-5.4}      & 10.9 & 21.7 & 16.3 & 24.6 \\
\texttt{GPT-5.4-mini}        & 10.4 & 20.9 & 15.8 & 23.7 \\
\texttt{GPT-5.3-chat}        & 8.7  & 18.2 & 13.9 & 20.4 \\
\bottomrule
\end{tabular}
\end{table}

\section{Case Studies}
\label{app:case_study}

We present qualitative comparisons of model outputs given identical input papers. For each case, we show the input prior works, followed by the outputs from \texttt{Qwen3-8B Base} and our best trained variant (\texttt{SFT}). The SFT model produces structured reasoning traces that explicitly synthesize across papers using citation handles, while the base model tends to produce more generic, list-based proposals.

\subsection{Case Study 1: Object-Centric World Models with Latent Actions}
\label{app:case_1}

\begin{headerbox}[Input: Prior Works $\mathcal{P}$]{goalblue}{goalbluebg}
\small
\begin{itemize}[leftmargin=*,itemsep=2pt]
\item \textbf{DDLP: Unsupervised object-centric video prediction with deep dynamic latent particles} (Tal Daniel et al., 2024). \textit{We propose a new object-centric video prediction algorithm based on the deep latent particle (DLP) representation. In comparison to existing slot- or patch-based representations, D...}
\item \textbf{Unsupervised image representation learning with deep latent particles} (Tal Daniel et al., 2022). \textit{We propose a new representation of visual data that disentangles object position from appearance. Our method, termed Deep Latent Particles (DLP), decomposes the visual input into l...}
\item \textbf{Genie: Generative interactive environments} (Jake Bruce et al., 2024). \textit{We introduce Genie, the first generative interactive environment trained in an unsupervised manner from unlabelled Internet videos. The model can be prompted to generate an endless...}
\item \textbf{PlaySlot: Learning Controllable Object-Centric Dynamics with Latent Actions} (Victor Villar-Corrales et al., 2025). \textit{Predicting future scene representations is a crucial task for enabling robots to understand and interact with the environment. However, most existing methods rely on videos and sim...}
\item \textbf{AdaWorld: Learning Adaptable World Models with Latent Actions} (Shenyuan Gao et al., 2025). \textit{World models aim to learn action-controlled future prediction and have proven essential for the development of intelligent agents. However, most existing world models rely heavily...}
\item \textbf{Improving generative imagination in object-centric world models} (Zhixuan Lin et al., 2020). \textit{The remarkable recent advances in object-centric generative world models raise a few questions. First, while many of the recent achievements are indispensable for making a general...}
\item \textbf{Entity-centric reinforcement learning for object manipulation from pixels} (Dan Haramati et al., 2024). \textit{Manipulating objects is a hallmark of human intelligence, and an important task in domains such as robotics. In principle, Reinforcement Learning (RL) offers a general approach to...}
\end{itemize}
\end{headerbox}

\vspace{4pt}

\begin{headerbox}[Output: \texttt{Qwen3-8B Base}]{neutralgray}{neutralgraybg}
\small
\textbf{Thinking trace (excerpt):}
\begin{quote}\itshape\small
Okay, let me try to work through this. The user provided several research papers and wants me to analyze them, find gaps, and propose a novel research idea. Let me start by understanding each paper's contribution and then see where the gaps might be.

First, the DDLP paper introduces a method for object-centric video prediction using deep dynamic latent particles. They use DLPs which are interpretable and efficient. The other DLP paper from 2022 is about unsupervised image representation learning, decomposing images into particles with spatial and feature info. Then there's Genie, a generative interactive environment trained unsupervised from internet videos, which can generate virtual worlds. PlaySlot is an object-centric model that uses unlabeled videos to infer object representations an [...]
\end{quote}
\textbf{Proposal:}
\begin{quote}\small
\textbf{Research Gap and Novel Research Idea}
\textbf{Gap:}
Current object-centric video prediction and world modeling approaches excel in representation learning, uncertainty estimation, and action inference but lack a unified framework that integrates \textbf{interpretability}, \textbf{temporal uncertainty}, \textbf{latent action control}, and \textbf{environmental adaptability}. Existing methods often prioritize one aspect (e.g., prediction accuracy or controllability) at the expense of others, limiting their applicability in dynamic, real-world scenarios requiring safe and interpretable planning. For example:
- \textbf{DDLP} and \textbf{DLP} provide interpretable object representations but lack robust uncertainty quantification for dynamic interactions.
- \textbf{Genie} and \textbf{PlaySlot} enable controllable environments but rely on labeled data or predefined action policies, limiting adaptability.
- \textbf{AdaWorld} and \textbf{G-SWM} address adaptability and uncertainty but struggle to integrate object-centric representations with actionable latent actions.
- \textbf{Entity-Centric RL} focuses on manipulation but does not explicitly model temporal uncertainty or generalize to novel environments.

\textbf{Novel Research Idea:}
\textbf{Uncertainty-Aware Adaptive Object-Centric World Models (UA-OCWM)}
\textbf{Objective:} Develop a unified framework that combines \textbf{interpretable object-centric representations} (DLPs), \textbf{temporal uncertainty estimation} (G-SWM), \textbf{latent action inference} (PlaySlot/AdaWorld), and \textbf{environmental adaptability} (AdaWorld) to enable \textbf{safe, controllable, and interpretable world modeling} for dynamic environments.

\textbf{Key Innovations:}
1. \textbf{Interpretable Uncertainty Estimation:}
   - Integrate DLPs with G-SWM's multimodal uncertainty framework to quantify uncertainty in object positions, sizes, and interactions during video prediction. This enables ``what-if'' scenarios (DDLP) while providing confidence bounds for future states.

2. \textbf{Adaptive Latent Action Control:}
   - Leverage AdaWorld's self-supervised latent action extraction and PlaySlot's action-conditioned prediction to infer and control actions in novel environments. This allows the model to adapt to unseen dynamics while maintaining interpretability.

3. \textbf{Dynamic Object Interaction Modeling:}
   - Extend DLPs to explicitly model spatial-temporal relationships between objects (e.g., occlusions, collisions) using a \textbf{graph-based particle interaction module}, enabling robust prediction in complex scenes.

4. \textbf{Efficient Transfer Learning:}
   - Pre-train the model on unlabeled videos (as in Genie) and fine-tune it on novel environments with minimal interaction, using uncertainty-aware meta-learning to adapt to heterogeneous actions.

\textbf{Applications:}
- \textbf{Robotics:} Safe manipulation tasks where uncertainty estimation ensures robustness to sensor noise and environmental changes.
- \textbf{Virtual Environments:} Interactive simulations with interpretable physics and controllable agen

[...]
\end{quote}
\end{headerbox}

\vspace{4pt}

\begin{headerbox}[Output: \texttt{Qwen3-8B SFT}]{guidepurple}{guidepurplebg}
\small
\textbf{Thinking trace (excerpt):}
\begin{quote}\itshape\small
I began from the observation that object-centric latent dynamics had already been made practical by Deep Latent Particles. @daniel2022dlp showed that a VAE can represent a scene as a set of spatially localized latent particles with position and appearance features, and that this representation supports both reconstruction and downstream control. The limitation for my purpose was that DLPs are primarily a representation, not a full world model, so they do not solve action-conditioned future prediction. @daniel2024ddlp extended DLPs to video prediction, demonstrating that particle trajectories can be modeled autoregressively and used for controllable generation, but it still did not provide a learned latent action space for control. In parallel, @bruce2024genie established that latent action [...]
\end{quote}
\textbf{Proposal:}
\begin{quote}\small
\textbf{[Motivation]}\\
Current world models face a specific technical tradeoff between object-centric structure and controllability. Object-centric methods identify and track entities, but they do not typically provide a learned latent action space that is both compact and directly usable for control. This limitation is visible in object-centric video prediction systems such as @daniel2024ddlp, which model particle trajectories but do not explicitly infer latent actions for controllable manipulation. It is also reflected in structured visual RL and world modeling work such as @haramati2024entitycentric and @lin2020gswm, which emphasize object identity and temporal imagination, yet still struggle to expose a representation that supports generalizable action-conditioned prediction.

At the same time, action-conditioned world models show that latent actions can be inferred from video dynamics without explicit action labels. @bruce2024genie demonstrates that temporal motion can be converted into a controllable latent action space, and @gao2025adaworld shows that such action-conditioned prediction can be learned from unlabeled video. However, these approaches operate on non-object-centric latent states, so they do not distinguish which entities should be acted upon or how actions should be organized across entities. This matters because manipulation and planning require actions to be attached to specific objects rather than to a monolithic latent state.

The resulting gap is therefore clear: there is no existing framework that combines a compact object-centric latent state with a learned latent action space that is explicitly aligned to those entities. This gap creates an opportunity to unify representation learning and action modeling in a single world model. If the action predictor is constrained to operate in the same particle space used by the world model, then the model can infer compact actions for controllable entities while preserving object identity across time. That formulation would directly address the limitations exposed by @villarcorrales2025playslot, @haramati2024entitycentric, and @lin2020gswm, while retaining the unsupervised action-learning advantages established by @bruce2024genie and @gao2025adaworld.

\par\medskip\noindent\textbf{[Method]}\par\smallskip
We propose a world model that represents each scene as a set of deep latent particles and learns a compact latent action space directly in that particle space. The core idea is to let the action predictor infer motion patterns from the current particle state, then condition future particle generation on those actions. This formulation addresses the identified gap by coupling object-centric state representation with object-centric action prediction, so the model can generate controllable futures without relying on hand-designed action dimensions or dense action annotations.

1. \textbf{Deep Latent Particle World Model}: We represent each frame as a set of spatially localized latent particles with position and appearance featur

[...]
\end{quote}
\end{headerbox}

\vspace{4pt}

\begin{headerbox}[Analysis]{rubricgreen}{rubricgreenbg}
\small
\textbf{Gap identification.} The SFT model identifies a precise technical gap: object-centric representations exist but lack a learned latent action space for control. The base model identifies a vague ``unified framework'' gap without specifying what is technically missing.

\textbf{Cross-paper synthesis.} The SFT model constructs an explicit logical chain: DLPs provide structure $\rightarrow$ latent actions from Genie/AdaWorld remove supervision bottleneck $\rightarrow$ combining both in a shared particle space enables entity-aligned control. The base model lists contributions independently without showing how they connect into a design.

\textbf{Methodological specificity.} The SFT model proposes five concrete components (particle-based dynamics, per-particle action predictor, entity-aligned control, unsupervised action extraction, generalization via particle-aligned actions), each justified by a specific prior work limitation. The base model proposes a generic ``Uncertainty-Aware Object-Centric World Model'' with a laundry list of applications and potential challenges but no clear technical mechanism.
\end{headerbox}

\subsection{Case Study 2: GFlowNet-Based Formulaic Alpha Factor Mining}
\label{app:case_2}

\begin{headerbox}[Input: Prior Works $\mathcal{P}$]{goalblue}{goalbluebg}
\small
\begin{itemize}[leftmargin=*,itemsep=2pt]
\item \textbf{Generating Synergistic Formulaic Alpha Collections via Reinforcement Learning} (Shuo Yu et al., 2023). \textit{In the field of quantitative trading, it is common practice to transform raw historical stock data into indicative signals for the market trend. Such signals are called alpha facto...}
\item \textbf{QuantFactor REINFORCE: Mining Steady Formulaic Alpha Factors with Variance-bounded REINFORCE} (Junjie Zhao et al., 2024). \textit{Alpha factor mining aims to discover investment signals from the historical financial market data, which can be used to predict asset returns and gain excess profits. Powerful deep...}
\item \textbf{Flow Network Based Generative Models for Non-Iterative Diverse Candidate Generation} (Emmanuel Bengio et al., 2021). \textit{This paper is about the problem of learning a stochastic policy for generating an object (like a molecular graph) from a sequence of actions, such that the probability of generatin...}
\item \textbf{Trajectory Balance: Improved Credit Assignment in GFlowNets} (Nikolay Malkin et al., 2022). \textit{Generative flow networks (GFlowNets) are a method for learning a stochastic policy for generating compositional objects, such as graphs or strings, from a given unnormalized densit...}
\item \textbf{Let the Flows Tell: Solving Graph Combinatorial Problems with GFlowNets} (Dinghuai Zhang et al., 2023). \textit{Combinatorial optimization (CO) problems are often NP-hard and thus out of reach for exact algorithms, making them a tempting domain to apply machine learning methods. The highly s...}
\item \textbf{Modeling Relational Data with Graph Convolutional Networks} (Michael Schlichtkrull et al., 2018). \textit{Knowledge graphs enable a wide variety of applications, including question answering and information retrieval. Despite the great effort invested in their creation and maintenance,...}
\item \textbf{AlphaForge: A Framework to Mine and Dynamically Combine Formulaic Alpha Factors} (Hao Shi et al., 2025). \textit{The complexity of financial data, characterized by its variability and low signal-to-noise ratio, necessitates advanced methods in quantitative investment that prioritize both perf...}
\end{itemize}
\end{headerbox}

\vspace{4pt}

\begin{headerbox}[Output: \texttt{Qwen3-8B Base}]{neutralgray}{neutralgraybg}
\small
\textbf{Thinking trace (excerpt):}
\begin{quote}\itshape\small
Okay, let me try to work through this. The user provided several research papers related to alpha factor mining and generative models, and they want me to analyze them, identify gaps, and propose a novel research idea.

First, I need to understand each paper's contribution. The first paper, Yu2023, talks about using reinforcement learning (RL) to mine synergistic alpha factors by optimizing the downstream combination model's performance. They use RL to explore the search space of formulaic alphas, which are interpretable and preferred in risk-sensitive markets. The second paper, Zhao2024, critiques PPO for alpha mining and proposes a modified REINFORCE algorithm with a dedicated baseline and information ratio as a reward. They mention improving correlation with returns by 3.83\% and better  [...]
\end{quote}
\textbf{Proposal:}
\begin{quote}\small
\textbf{Novel Research Idea: Synergistic Alpha Generation with Graph-Enhanced GFlowNets and Adaptive Reinforcement Learning}

\textbf{Motivation and Context}
The existing literature on alpha factor mining and combination (Yu2023, Zhao2024, Shi2025) emphasizes interpretability, dynamic weighting, and reinforcement learning (RL) for optimizing alpha sets. However, critical gaps remain:
1. \textbf{Limited Diversity in Alpha Generation}: While GFlowNets (Bengio2021, Zhang2023) excel at generating diverse solutions, they are not explicitly tailored for financial alphas, which require balancing interpretability and market adaptability.
2. \textbf{Decoupled Generation and Combination}: Most frameworks (Yu2023, Shi2025) treat alpha generation and combination as separate stages, missing opportunities for end-to-end optimization.
3. \textbf{Neglect of Relational Market Dynamics}: Financial markets involve complex relationships (e.g., sector correlations, sentiment networks), which are underexplored in alpha generation.
4. \textbf{Temporal Adaptability}: Current methods (Zhao2024, Shi2025) rely on static or semi-static weighting, failing to dynamically adjust to rapidly shifting market conditions.

\textbf{Proposed Framework: Graph-Enhanced GFlowNets with Adaptive Reinforcement Learning (GE-GFlowRL)}
\textbf{Core Idea}: Integrate \textbf{GFlowNets} for diverse alpha generation, \textbf{graph-based relational modeling} (e.g., R-GCNs) to capture market dependencies, and \textbf{adaptive RL} for dynamic combination, creating a unified framework that addresses the above gaps.

\textbf{Key Components}
1. \textbf{Graph-Enhanced Alpha Generation (GFlowNets + R-GCNs)}
   - \textbf{GFlowNets}: Generate a diverse set of formulaic alphas by modeling the generation process as a flow network, ensuring coverage of high-reward (high-accuracy) and low-reward (diverse) solutions.
   - \textbf{R-GCNs}: Encode relational data (e.g., market entities, sector relationships, sentiment networks) to inform alpha generation. For example, R-GCNs can model how macroeconomic factors (nodes) influence stock returns (edges), enabling alphas that capture these interactions.
   - \textbf{Hybrid Objective}: Combine GFlowNet's flow consistency with R-GCN's graph-based rewards to prioritize alphas that are both high-performing and interpretable (e.g., leveraging graph edges as feature interactions).

2. \textbf{Adaptive Reinforcement Learning for Dynamic Combination}
   - \textbf{State Representation}: Use a graph embedding of the current market state (e.g., via R-GCNs) to contextualize alpha combinations.
   - \textbf{Policy Optimization}: Train a policy (e.g., PPO or modified REINFORCE) to dynamically adjust weights of generated alphas, with rewards based on:
     - \textbf{Short-term Performance}: Recent portfolio returns.
     - \textbf{Risk-adjusted Metrics}: Sharpe ratio, Value at Risk (VaR), or information ratio.
     - \textbf{Diversity Constraints}: Encourage exploration of novel alpha combinations to avoid overfitting to historical patterns.

[...]
\end{quote}
\end{headerbox}

\vspace{4pt}

\begin{headerbox}[Output: \texttt{Qwen3-8B SFT}]{guidepurple}{guidepurplebg}
\small
\textbf{Thinking trace (excerpt):}
\begin{quote}\itshape\small
I began from the observation that reinforcement-learning-based alpha mining had already established a strong search paradigm, but only for single-factor discovery. @yu2023alphacollections showed that a policy can generate formulaic alphas and optimize them through downstream portfolio returns, which made RL the natural optimization backbone for alpha generation. However, its objective still depends on evaluating complete factor sequences, so the search remains expensive. @zhao2024quantfactor reinforced this point by showing that even improved RL training objectives for alpha mining still fundamentally rely on trajectory-level reward signals, exposing the inefficiency of evaluating entire factor sequences during training. In parallel, @bengio2021gflownet provided a different principle for s [...]
\end{quote}
\textbf{Proposal:}
\begin{quote}\small
\textbf{[Motivation]}\\
Formulaic alpha factor mining requires discovering sequences of interpretable formulas whose combined portfolio return is maximized. Existing RL-based approaches have shown that this objective can indeed be optimized effectively: @yu2023alphacollections demonstrates that reinforcement learning can generate synergistic alpha sets by optimizing downstream portfolio returns, and @zhao2024quantfactor further shows that improved RL training can make alpha mining more stable. However, both lines of work still evaluate factor sequences only at the terminal stage and optimize return through trajectory-level rewards. This creates a concrete inefficiency: each generated factor sequence must be fully realized before its quality can be assessed, making training substantially more expensive than the trajectory-level evaluation used in standard RL.

A second limitation is that current alpha generators do not adequately model the structural dependencies among factors during generation. @shi2025alphaforge highlights the importance of diversity in factor discovery, but its search procedure still does not explicitly represent the compositional structure of factor sequences in the policy. Meanwhile, @schlichtkrull2018rgcn shows that graph-based encoders can reason over relational structure, suggesting an opportunity to inject structural information into alpha generation. This gap matters because alpha quality is not determined by isolated formulas; it depends on how formulas interact across time and how the resulting factor sequence composes into a coherent trading signal.

These observations create a clear opportunity. GFlowNets provide a principled alternative to return maximization by learning a policy whose terminal distribution matches an unnormalized reward distribution, as proposed in @bengio2021gflownet, and trajectory balance provides a stable training objective for long-horizon sequential generation in @malkin2022trajectorybalance. If alpha factor mining can be reformulated as sampling factor sequences whose probability is proportional to their portfolio return, then the search can be guided by a more informative intermediate signal than the final reward alone. The central problem is therefore to design a GFlowNet-based alpha generator that approximates the full return distribution efficiently while incorporating structural information about factor sequences.

\par\medskip\noindent\textbf{[Method]}\par\smallskip
We propose to cast alpha factor mining as a generative flow problem in which the policy learns to sample factor sequences whose terminal distribution is proportional to their portfolio return. Instead of maximizing terminal reward directly, the model approximates the unnormalized reward with a discounted sum of immediate returns, allowing early actions to receive informative reward estimates during training. This formulation addresses the inefficiency of trajectory-level RL while preserving the objective of maximizing portfolio return.

1. \textbf{GFlowNet formulation}:

[...]
\end{quote}
\end{headerbox}

\vspace{4pt}

\begin{headerbox}[Analysis]{rubricgreen}{rubricgreenbg}
\small
\textbf{Gap identification.} The SFT model identifies a precise inefficiency: trajectory-level reward evaluation in RL-based alpha mining is expensive because full factor sequences must be realized before quality assessment. The base model identifies a broad ``decoupled generation and combination'' gap without specifying why this matters computationally.

\textbf{Cross-paper synthesis.} The SFT model builds a logical argument: RL alpha mining works but evaluates only at trajectory end $\rightarrow$ GFlowNets replace return maximization with distribution matching $\rightarrow$ trajectory balance stabilizes long-horizon generation $\rightarrow$ R-GCN supplies structural context for intermediate reward estimation. The base model proposes combining GFlowNets + R-GCNs + RL but does not explain the causal logic connecting these choices.

\textbf{Methodological specificity.} The SFT model proposes a concrete mechanism: GFlowNet over factor sequences with discounted immediate-return approximation and R-GCN conditioning. Each component has a clear role (what it does, why it is needed, which prior work gap it addresses). The base model enumerates ``key components'' but presents them as a feature list without a unifying technical mechanism.
\end{headerbox}

\subsection{Case Study 3: Safety-Aware Low-Rank Adaptation}
\label{app:case_3}

\begin{headerbox}[Input: Prior Works $\mathcal{P}$]{goalblue}{goalbluebg}
\small
\begin{itemize}[leftmargin=*,itemsep=2pt]
\item \textbf{LoRA: Low-Rank Adaptation of Large Language Models} (Hu et al., 2022). \textit{An important paradigm of natural language processing consists of large-scale pre-training on general domain data and adaptation to particular tasks or domains. As we pre-train larg...}
\item \textbf{Safety-Aligned LoRA (SaLoRA): Safety-Alignment Preserved Low-Rank Adaptation} (Li et al., 2025). \textit{As advancements in large language models (LLMs) continue and the demand for personalized models increases, parameter-efficient fine-tuning (PEFT) methods (e.g., LoRA) will become e...}
\item \textbf{Safe LoRA: The Silver Lining of Reducing Safety Risks When Finetuning Large Language Models} (Hsu et al., 2024). \textit{While large language models (LLMs) such as Llama-2 or GPT-4 have shown impressive zero-shot performance, fine-tuning is still necessary to enhance their performance for customized...}
\item \textbf{Fine-tuning Aligned Language Models Compromises Safety, Even When Users Do Not Intend To!} (Qi et al., 2024). \textit{Optimizing large language models (LLMs) for downstream use cases often involves the customization of pre-trained LLMs through further fine-tuning. Meta's open release of Llama mode...}
\item \textbf{AlphaEdit: Null-Space Constrained Model Editing for Language Models} (Fang et al., 2025). \textit{Large language models (LLMs) often exhibit hallucinations due to incorrect or outdated knowledge. Hence, model editing methods have emerged to enable targeted knowledge updates. To...}
\item \textbf{PiSSA: Principal Singular Values and Singular Vectors Adaptation of Large Language Models} (Meng et al., 2024). \textit{To parameter-efficiently fine-tune (PEFT) large language models (LLMs), the low-rank adaptation (LoRA) method approximates the model changes $\Delta W \in \mathbb{R}^{m \times n}$...}
\item \textbf{Intrinsic Dimensionality Explains the Effectiveness of Language Model Fine-Tuning} (Aghajanyan et al., 2021). \textit{Although pretrained language models can be fine-tuned to produce state-of-the-art results for a very wide range of language understanding tasks, the dynamics of this process are no...}
\end{itemize}
\end{headerbox}

\vspace{4pt}

\begin{headerbox}[Output: \texttt{Qwen3-8B Base}]{neutralgray}{neutralgraybg}
\small
\textbf{Thinking trace (excerpt):}
\begin{quote}\itshape\small
Okay, let's see. The user provided several research papers related to fine-tuning large language models (LLMs) with a focus on parameter-efficient methods like LoRA and its variants. They also mentioned issues around safety alignment and the effectiveness of these methods. The task is to analyze these papers, identify gaps, and propose a novel research idea.

First, I need to understand each paper's contribution. LoRA is about reducing trainable parameters by using low-rank matrices. Then SaLoRA and SafeLoRA address safety alignment issues during fine-tuning, showing that even benign data can affect safety. AlphaEdit is about model editing without disrupting existing knowledge. PiSSA improves upon LoRA by using principal components for faster convergence. The intrinsic dimensionality paper [...]
\end{quote}
\textbf{Proposal:}
\begin{quote}\small
Novel Research Idea: \textbf{Intrinsic-Dimension-Guided Safety-Enhanced Low-Rank Adaptation (ID-SaLoRA)}

\textbf{Motivation \& Context}
Recent advancements in parameter-efficient fine-tuning (PEFT) methods like \textbf{LoRA} have enabled efficient adaptation of large language models (LLMs) while preserving most of the pre-trained knowledge. However, safety alignment risks persist, as demonstrated by studies showing that even benign fine-tuning can degrade safety (e.g., \textbf{Qi et al. 2024}, \textbf{SafeLoRA 2024}). Meanwhile, the \textbf{intrinsic dimensionality} of LLMs (Aghajanyan et al. 2021) reveals that only a small subset of parameters is critical for performance, suggesting that safety and efficiency could be co-optimized within this low-dimensional subspace.

\textbf{Research Gap}
Current safety-enhanced PEFT methods (e.g., \textbf{SaLoRA}, \textbf{SafeLoRA}) apply constraints globally or across the entire model, potentially sacrificing efficiency or performance. However, the \textbf{intrinsic dimensionality} framework suggests that safety-critical components may reside in a low-dimensional subspace, while other parameters are redundant. This gap highlights the need for a method that:
1. \textbf{Leverages intrinsic dimensionality} to identify safety-critical parameters.
2. \textbf{Applies safety constraints selectively} to this subspace, preserving efficiency.
3. \textbf{Maintains performance} on downstream tasks by avoiding disruption of non-critical parameters.

\textbf{Proposed Approach}
\textbf{ID-SaLoRA} combines \textbf{intrinsic dimensionality analysis} with \textbf{safety-preserving PEFT} (e.g., SaLoRA) to achieve efficient and safe adaptation:
1. \textbf{Intrinsic Subspace Identification}:
   - Use singular value decomposition (SVD) or random projection to identify the low-dimensional subspace (as in \textbf{PiSSA} and \textbf{Aghajanyan et al.}) that captures most of the model's performance.
   - This subspace corresponds to the ``intrinsic'' parameters critical for task performance and safety alignment.

2. \textbf{Safety-Enhanced Low-Rank Adaptation}:
   - Apply \textbf{SaLoRA}-style safety constraints \textbf{only to the intrinsic subspace}, freezing or regularizing parameters outside this subspace.
   - This ensures that safety alignment is preserved without over-constraining the model.

3. \textbf{Efficient Training}:
   - Use \textbf{low-rank decomposition} (as in LoRA) to adapt only the intrinsic subspace, reducing trainable parameters by orders of magnitude.
   - Combine with \textbf{quantization} (e.g., QPiSSA) to further reduce memory usage.

4. \textbf{Theoretical Justification}:
   - Leverage the \textbf{intrinsic dimensionality} framework to derive generalization bounds that are independent of full parameter counts, ensuring safety and efficiency are aligned.

\textbf{Expected Contributions}
1. \textbf{First integration of intrinsic dimensionality with safety-aware PEFT}, enabling targeted safety constraints without sacrificing efficiency.
2. \textbf{Empirical validation} on safety benchmarks (e.g., ja

[...]
\end{quote}
\end{headerbox}

\vspace{4pt}

\begin{headerbox}[Output: \texttt{Qwen3-8B SFT}]{guidepurple}{guidepurplebg}
\small
\textbf{Thinking trace (excerpt):}
\begin{quote}\itshape\small
I started from the observation that LoRA already provides a strong structural handle on adaptation: it freezes the backbone and learns a low-rank update, so any safety intervention can be inserted directly into that update space rather than into full fine-tuning @hu2022lora. The limitation is that the update space is unconstrained, and prior safety work had already shown that even benign downstream adaptation can degrade alignment @qi2024finetuning. SafeLoRA addressed this by projecting the LoRA update into a safety-aligned subspace using a reference model @hsu2024safelora, but its projection is a hard constraint that can suppress task learning. SaLoRA improved that formulation by replacing the safety subspace with a null space, which preserves more task capacity @li2025salora, yet it stil [...]
\end{quote}
\textbf{Proposal:}
\begin{quote}\small
\textbf{[Motivation]}\\
Fine-tuning aligned large language models with LoRA is attractive because it preserves the backbone and keeps adaptation lightweight, but recent evidence shows that this efficiency comes with a safety cost. Even benign downstream adaptation can degrade safety alignment, indicating that the LoRA update space can drift in harmful directions during optimization @qi2024finetuning. Existing safety-aware LoRA methods address this by constraining the update, but they do so through fixed projection operators. SafeLoRA projects the update into a safety-aligned subspace defined by reference models @hsu2024safelora, while SaLoRA uses a null-space formulation to preserve more task capacity @li2025salora. Both methods reveal the same structural limitation: safety is enforced by a static geometric constraint rather than by an estimate derived from the current task and model state.

This gap matters because hard constraints can be too rigid for downstream learning. If the task update already contains safety-preserving information, then forcing it into a fixed null space or a fixed safety subspace can remove exactly the directions needed for performance. The opportunity, suggested by @meng2024pissa, is to treat the LoRA update geometry itself as the object of analysis: if safety-relevant directions can be identified from the update space, then safety preservation can be enforced through a data-dependent constraint rather than a hand-designed one. At the same time, @aghajanyan2021intrinsic suggests that fine-tuning occupies a low-dimensional manifold, making it plausible that safety alignment can also be represented by a compact subspace. The problem we therefore address is: how can we estimate a task-specific safety subspace from the model's own behavior, and how can we use that estimate to constrain LoRA updates so that safety is preserved without sacrificing downstream utility?

We define the research problem as constructing a safety-aware LoRA method that is adaptive to the downstream task while remaining lightweight and compatible with frozen backbones. The key technical challenge is to estimate a safety-relevant subspace from limited safety-aligned data and to convert that estimate into a trainable update constraint that suppresses harmful directions while leaving task-relevant directions available for learning.

\par\medskip\noindent\textbf{[Method]}\par\smallskip
We propose Safety LoRA, a safety-aware fine-tuning framework that estimates a safety subspace from the model's own safety-aligned outputs and constrains the downstream LoRA update to remain in the null space of that subspace. Instead of projecting the update into a fixed safety-aligned or null space as in prior methods @hsu2024safelora @li2025salora, we estimate the dominant safety directions through SVD and remove only the most influential safety-irrelevant directions. This formulation directly addresses the rigidity of fixed constraints while preserving the low-rank adaptation efficiency introduced by @hu2022lo

[...]
\end{quote}
\end{headerbox}

\vspace{4pt}

\begin{headerbox}[Analysis]{rubricgreen}{rubricgreenbg}
\small
\textbf{Gap identification.} The SFT model identifies a specific failure mode: fixed safety projection constraints can over-regularize because they treat safety as a static property rather than a task-conditioned geometric structure. The base model identifies a broader ``global vs.\ selective constraint'' gap without pinpointing why existing constraints fail.

\textbf{Cross-paper synthesis.} The SFT model traces a precise design path: LoRA provides the update space $\rightarrow$ safety degradation persists even under benign adaptation $\rightarrow$ existing projections (SafeLoRA, SaLoRA) are too rigid $\rightarrow$ PiSSA shows SVD can identify important directions $\rightarrow$ AlphaEdit shows null-space projection preserves behaviors $\rightarrow$ combining these yields data-dependent safety estimation. The base model mentions similar papers but does not construct a causal chain showing why their combination produces a specific new mechanism.

\textbf{Methodological specificity.} The SFT model proposes four tightly coupled steps: estimate safety subspace from model outputs, extract dominant directions via SVD, project LoRA updates into the null space of those directions, and train under this adaptive constraint. The base model proposes ``ID-SaLoRA'' but describes it at a higher level of abstraction without specifying how intrinsic dimensionality analysis concretely feeds into the safety constraint.
\end{headerbox}

\newpage